\documentclass[12pt]{article}

\usepackage{arxiv}

\usepackage[utf8]{inputenc} % allow utf-8 input
\usepackage[T1]{fontenc}    % use 8-bit T1 fonts
\usepackage{hyperref}       % hyperlinks
\usepackage{url}            % simple URL typesetting
\usepackage{booktabs}       % professional-quality tables
\usepackage{amsfonts}       % blackboard math symbols
\usepackage{nicefrac}       % compact symbols for 1/2, etc.
\usepackage{microtype}      % microtypography
\usepackage{lipsum}
\usepackage{fancyhdr}       % header
\usepackage{graphicx}       % graphics
\usepackage{subcaption}
 \usepackage[greek,english]{babel} 
 \usepackage[artemisia]{textgreek}  
 
\usepackage{doi}

\title{Golyadkin's torment: \dopps\ and Adversarial Vulnerability} 

\date{} 					% Or removing it

\author{ \href{https://orcid.org/0000-0003-4560-4900}{\includegraphics[scale=0.06]{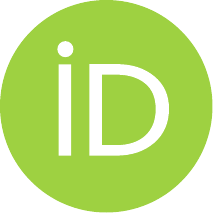}\hspace{1mm}George I. Kamberov}\thanks{	\texttt{george.i.kamberov@gmail.com}} \\
 University of Alaska Anchorage\\
	Anchorage, AK 99508 \\
	\texttt{gkamberov@alaska.edu} 
}

\renewcommand{\headeright}{Technical Report}
\renewcommand{\undertitle}{Technical Report}
\renewcommand{\shorttitle}{\textit{Golyadkin's torment}}

\hypersetup{
pdftitle={Golyadkin's torment: Doppelgangers and Adversarial Vulnerability},
pdfsubject={q-bio.NC, q-bio.QM},
pdfauthor={George I.~Kamberov},
pdfkeywords={Metamers \and Adversarial examples \and Active Discrimination},
}

\usepackage{orcidlink}
\usepackage{amsmath,amsfonts,amsthm,bm,amssymb,dsfont}
\newcommand{\Ups}{\Upsilon(\Phi)} %sigma-alg of feature reps

\newcommand{\charf}[1]{\mathop{\mathds{1}_{#1}}}%characteristic function indicator
\newcommand{\proto}[2]{\mathop{P}\left(#1,#2\right)}
\newcommand{\N}{\mathbb{N}}
\def\taup{\tau_{\textgreek{d}}} %perceptual topology
\newcommand{\card}[1]{\mathop{\mbox{\rm card}}\left(#1\right)}
\newcommand{\pind}{\stackrel{\textgreek{ad}}{\approx}}
 
 \newcommand{\picoco}{\stackrel{(c, \epsilon)}{\approx}}
\newcommand{\sorites}{\sim_{\sigma}} %the transitive closure 
\newcommand{\pgraph}{\Gamma\left(\X, E_{\alpha\delta}\right)}

\newcommand{\rgraph}{\Gamma\left(\X, E_{\approx}\right)} %relationship graph
\newcommand{\rcliques}{\mathop{\frak Cl}\left(\rgraph \right)}  %gen cliques

\newcommand{\clu}[1]{\mathop{\frak cl}(#1)}
\newcommand{\pdist}[2]{\mathop{{\it d}_w}\left(#1, #2\right)} %perceptual distance
\newcommand{\pdistNA}{\mathop{{\it d}_w}} %perceptual distance func
\newcommand{\epdist}[2]{\mathop{{\it d}_\infty}\left(#1, #2\right)} %extended perceptual distance
\newcommand{\epdistNA}{\mathop{{\it d}_\infty}} %extended perceptual distance func
\newcommand{\labelR}[1]{\mathop{\mbox{\rm label}_{R}}\left(#1\right)} %label()
\newcommand{\labelf}[1]{\mathop{\mbox{\rm label}}_{#1}} %label_
\newcommand{\classif}[2]{#1 = \{#1_1,\ldots,#1_{#2}\}} %classsifier(!, #2 elems
\newcommand{\classof}[2]{#1_{i(#2)}}
\newcommand{\X}{\mathbf{X}}

\newcommand{\erf}[1]{\mathop{\mbox{erf}}(#1)}   %error function

\newcommand{\xmod}{\X/\!\!\sim_\sigma}
\newcommand{\LLP}{satisfies the  Law of Indiscriminability}
\newcommand{\dfr}{discriminative feature representation}
\newcommand{\dfrs}{discriminative feature representations}
\newcommand{\Dfrs}{Discriminative feature representations}
\newcommand{\CA}{conceptually ambiguous}
\newcommand{\dopp}{Doppelg{\"a}nger}
\newcommand{\dopps}{Doppelg{\"a}ngers} 
\newcommand{\twins}[1]{\mathop{\mathfrak{d}}(#1)}
\newcommand{\adv}{adversarial}
\newcommand{\aE}{adversarial example}
\newcommand{\aEs}{adversarial examples}
\newcommand{\ML}{machine learning}
\newcommand{\adiakrisia}{ indiscriminability}
\newcommand{\adiak}{indiscriminable}
\newcommand{\plap}{\triangle_{\alpha\delta}}
\newcommand{\slap}{\triangle_{\sigma}}
\newcommand{\pdeg}{d_{\alpha\delta}}
\newcommand{\sdeg}{d_{\sigma}}
\def\F{\mathcal{F}}  %sigma algebra

\newcounter{lemma_counter}
\newcounter{theo_counter}
\newcounter{obs_counter}
\newcounter{e_counter}
\newtheorem{lem}[lemma_counter]{Lemma}
\newtheorem{theo}[theo_counter]{Theorem}
\newtheorem{obs}[obs_counter]{Observation}
\theoremstyle{definition}
\newtheorem{definition}{Definition}
\newcommand{\myeg}[1]{\refstepcounter{e_counter}   \label{eg:#1} \noindent {\bf Example  \ref{eg:#1}:} }
\newcommand{\newterm}[1]{{\bf #1}}

\def\eqref#1{equation~\ref{#1}}
\def\1{\bm{1}}

\DeclareMathAlphabet{\mathsfit}{\encodingdefault}{\sfdefault}{m}{sl}
\SetMathAlphabet{\mathsfit}{bold}{\encodingdefault}{\sfdefault}{bx}{n}

\newcommand{\R}{\mathbb{R}}

\begin{document}

% ---------------------------------------------------------------

\maketitle

\begin{abstract}
 Many machine learning (ML) classifiers are claimed to outperform humans, but they still make mistakes that humans do not. The most notorious examples of such mistakes are adversarial visual metamers. This paper aims to define and investigate the phenomenon of adversarial \dopps\ (AD),  which includes adversarial visual metamers, and to compare the performance and robustness of ML classifiers to human performance.  
 
 We find that AD are inputs that are close to each other with respect to a perceptual  metric defined in this paper, and show that AD are  qualitatively different from the usual adversarial examples. The vast majority of classifiers are vulnerable to  AD and  robustness-accuracy trade-offs may not improve them.  Some classification problems may not admit any AD robust classifiers because the underlying classes are ambiguous.  We provide criteria that can be used to determine whether a classification problem is well defined or not;  describe of an AD robust classifiers' structure and attributes; introduce and explore the notions of conceptual entropy and regions of conceptual ambiguity for classifiers that are vulnerable to AD attacks, along with methods to bound the AD fooling rate of an attack. We define the notion of classifiers that exhibit hyper-sensitive behavior, that is, classifiers whose only mistakes are adversarial \dopps.  Improving the AD robustness of hyper-sensitive classifiers is equivalent to improving accuracy. We identify conditions guaranteeing that all classifiers with sufficiently high accuracy are hyper-sensitive. 
 
Our findings aim at significant improvements in the reliability and security of machine learning systems.

\end{abstract}

\section{Introduction}
\label{sec:intro}
\begin{figure}[h]
\begin{center}
\begin{subfigure}{0.3\textwidth}
  \centering
  \includegraphics[width=0.7\linewidth]{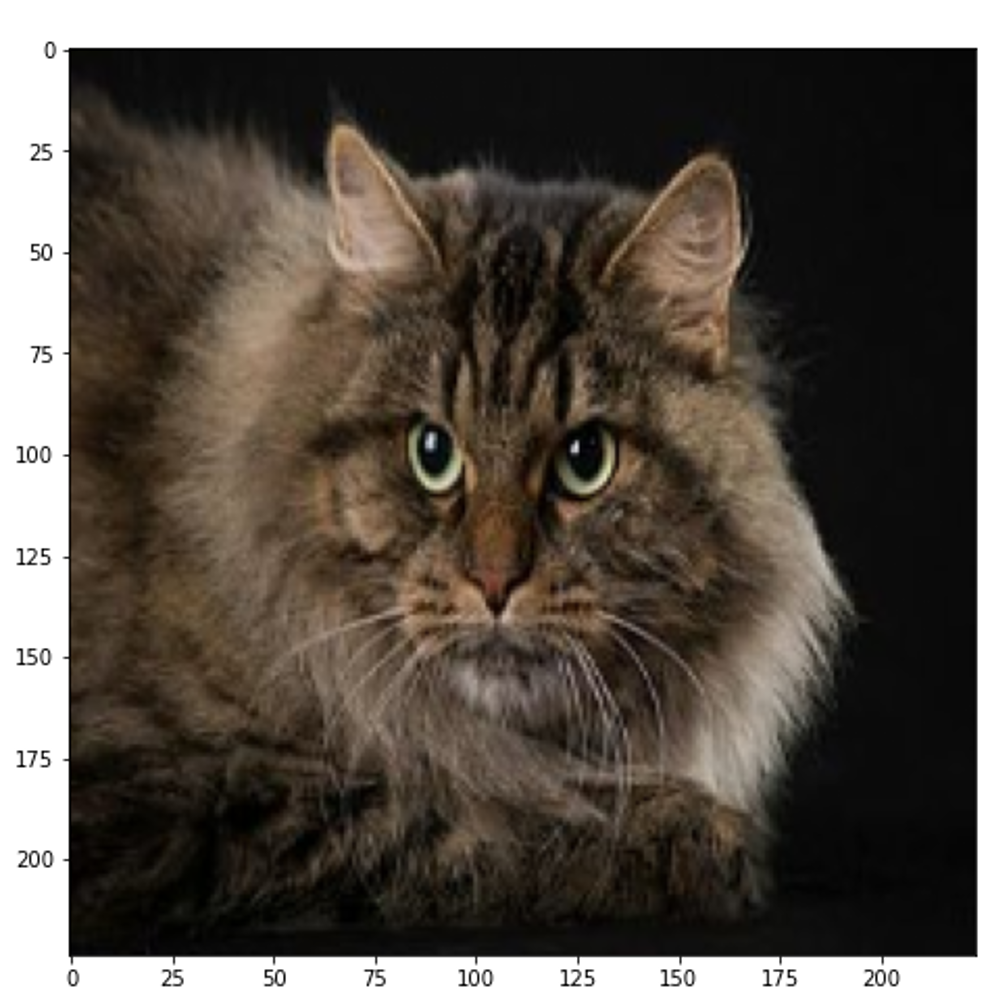}
  \caption{\ }
  \label{fig:sfig1}
\end{subfigure}%\hfil
\begin{subfigure}{0.3\textwidth}
  \centering
  \includegraphics[width=0.7\linewidth]{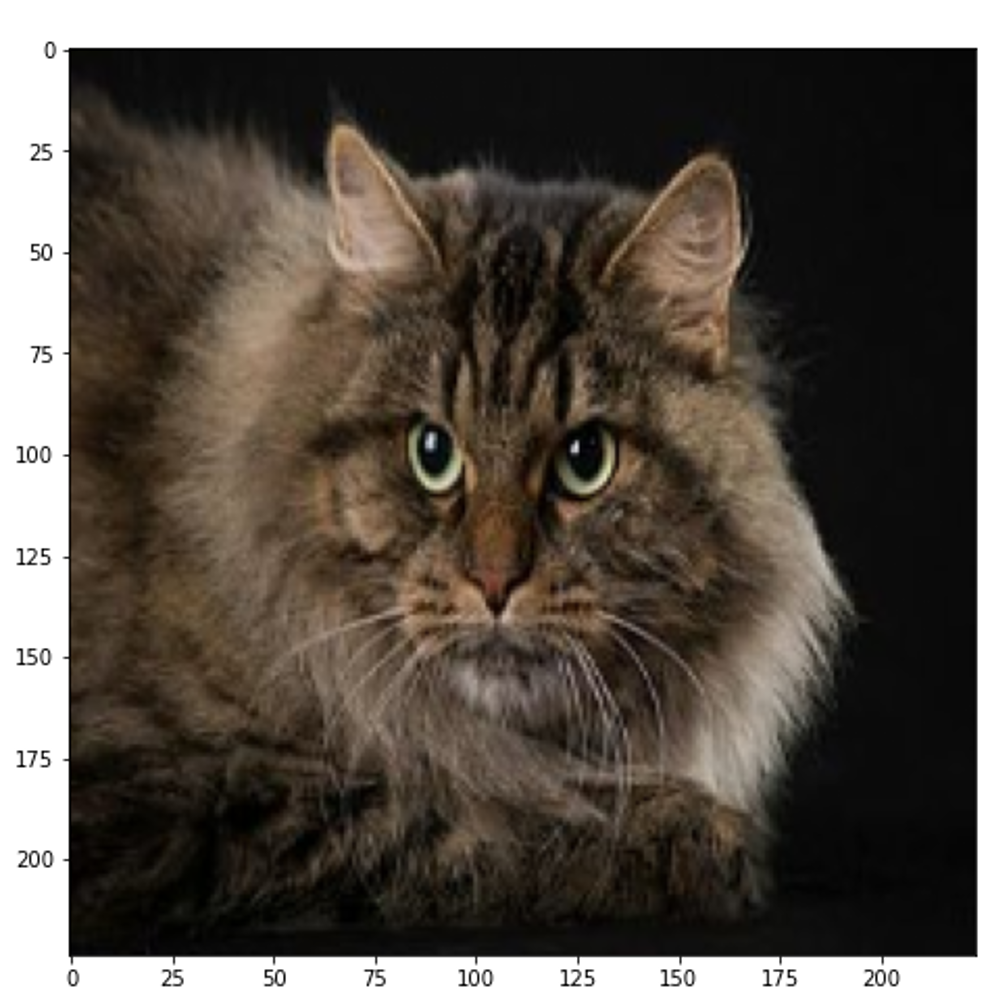}
  \caption{\ }
  \label{fig:sfig2}
\end{subfigure}
\hfil
\caption{Most people cannot discriminate image (a) from image (b).  MobileNetV2 classifies the later image as ``persian" and the former picture as ``taby".}  %where $0<|\delta_1|_{\ell_p}=|\delta_2|_{\ell_p}$}
\label{fig:metamers}
\end{center}
\end{figure}

\begin{figure}[h]
\begin{center}
%\end{subfigure}\\
\begin{subfigure}{.2\textwidth}
  \centering
  \includegraphics[width=\linewidth]{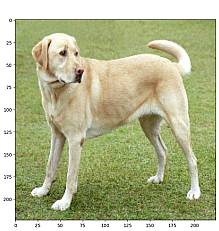}
  \caption{Labrador}
%  \label{fig:sfig1}
\end{subfigure}%\hfil
\begin{subfigure}{.2\textwidth}
  \centering
  \includegraphics[width=\linewidth]{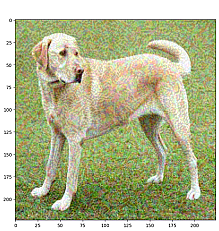}
  \caption{Weimeraner }
%  \label{fig:sfig2}
\end{subfigure}
\hfil
\caption{Applying a Fast Signed Gradient perturbation to the image (a) classified by MobileNetV2 as Labrador yields the image (b) which is  classified by  MobileNetV2 as Weimeraner.}  %where $0<|\delta_1|_{\ell_p}=|\delta_2|_{\ell_p}$}
\label{fig:dopps}
\end{center}
\end{figure}

Perceptual metamers\footnote{``images that are physically distinct but perceptually indistinguishable", \cite{broderick2023foveated}. See also ``metameric images", \cite{jagadeesh2022texture}}  are  the most striking adversarial examples studied by the machine learning community.  Two perceptual metamers are shown in  Figure \ref{fig:metamers}. The phenomenon of metamersim studied in the visual domain, including perceptual metamers, is  a manifestation of the  existence of \dopps: different  inputs or stimuli  that are perceptually indiscriminable.  
The research community has engaged in active studies of the adversarial   vulnerability of seemingly successful classifiers ever since the publication of \cite{szegedy2013intriguing}. Adversarial \dopps, that is, \aEs\ which are \dopps\ are   qualitatively different from the vast majority of  known \aEs \  which humans readily discriminate from correctly classified  input samples (Figure \ref{fig:dopps}).  
However, there is no evidence that  \dopps \  can be studied and  understood completely using   the $\ell_p$ norms or more general geodesic distances on manifolds that have been employed to quantify sample differences and to investigate \aEs. Perception and context impose topology on a space of inputs. We will denote this context-relative topology by $\tau_{\textgreek{d}}$.\footnote{The \textgreek{d} in $\tau_{\textgreek{d}}$ is just the first letter in the Greek word for discrimination ( \textgreek{di\'a}\textkappa\textgreek{rish}).}
It is defined by the context-relative ability to acquire and deploy knowledge. Whether or not it is always a manifold topology is an open question. 

 In this paper, we initiate a focused investigation of  context-relative perceptual topologies on a space of inputs  $\X$  and the vulnerability and robustness  of \ML\  classifiers to adversarial \dopps.  We show that, while the majority of classifiers are vulnerable to adversarial \dopps, safe (\dopp\ robust) classifiers do exist if the classification problem is well defined. However, they may be very rare. 
 
 In Section \ref{sec:indi}, we discuss the context-relative notion of (active) indiscriminability, the  topology $\taup$ that it induces on a space of inputs, various motivating examples of $\taup$, their separability and metrizability properties, and the relation between indiscernability, indsicriminability, and  feature representations.  The existence and structure of  \dopp\ robust classifiers is discussed in Section \ref{sec:Class}. 
In Section \ref{section:accuracy}, we investigate the relationship between \dopps \ and misclassified input samples, define the notion of hypersensitive behavior, and show that improved \adv\  \dopp \  robustness does not have to lead to a reduction in accuracy. In Section \ref{sec:distance}, we show that the distances between \dopps\ are small  if measured by a perceptually-based context-relative metric. Thus, adversarial \dopps\ are small perturbations.  

The attributes and structure of adversarially robust classifiers are discussed in  Section \ref{sec:similarity}. By definition, a classifier is not regular if and only if some inputs can be attacked by adversarial \dopps. Not surprisingly, it turns out that some inputs are more vulnerable than others. In Section \ref{sec:aa}, 
we provide measures of adversarial \dopps\ vulnerability and  upper bounds on the fooling rate of an adversarial \dopp\ attack. 

\section{Related work}\label{sec:related}
The pair $(\X, \mbox{indiscriminability relation})$ is a tolerance space. Tolerance spaces, rough sets, and granular computing have bee discussed  extensively. See  \cite{zeeman1962topology,Zeeman1965Topology,poston1971fuzzy,roberts1973tolerance,schreider1975equality,
sossinsky1986tolerance,hovsepian1992metalogical,yao1996two,henry2011near,peters2012tolerance,pawlak2012rough,zeeman2017tolerance,yao2019granularity}
The color and image  metamers studied by many authors including 
\cite{ostwald1919physikalische,wyszecki1953valenzmetrische,thornton1973matching,krantz1975color,cohen1985color,
balas2009summary,freeman2011metamers,logvinenko2013object,
Logvinenko2016Counting,feather2019metamers,jagadeesh2022texture,broderick2023foveated} are \dopps.\footnote{Some metamers arising in other fields including biology and chemistry are not \dopps, for example,  segments in many earthworms are considered metamers but are visually discriminable.}

 The research on \aEs\  to date builds on the hypothesis that  the space of input samples  is a metric space $\left(\X, \mbox{dist}_\X\right)$. A misclassified input $x^*$ is considered an \aE\  if it is nearby a correctly classified input sample $x$, i.e., $\mbox{dist}_\X (x, x^*)$ is small.  In fact usually $\X$ is  assumed to be $\R^n$, endowed with the  $\ell_p$, norm, $p=1,2,\ldots, \infty$ or at least locally homeomorphic to $\R^n$, i.e., a manifold, equipped with some geodesic distance.   
 Somewhat non surprisingly many authors have shown that every classifier can be attacked with such \aEs\  \cite{goodfellow2014explaining,papernot2016limitations,fawzi2018adversarial,boucher2022bad}
or at least that this is true in many contexts, \cite{moosavi2017universal,shafahi2018adversarial,mahloujifar2019curse}.  

 Other papers indicate that there are paths toward eliminating
\aEs\  completely, i.e., it is possible to achieve provable ``adversarial
robustness" by fixing/retraining the classifier
\cite{globerson2006nightmare,tanay2016boundary,madry2017towards,ilyas2019adversarial,sulam2020adversarial,allen2022feature}. 
A widely accepted tenet is that  ``there is a clear trade-off between accuracy and [adversarial] robustness, and a better performance in testing accuracy in general reduces [adversarial] robustness",  \cite{su2018robustness}. For empirical evidence for this trade-off and some attempts to explain this phenomenon see \cite{tsipras2018robustness,su2018robustness,zhang2019theoretically}. 

\section{Perceptual Topology} \label{sec:indi}
\subsection{Indiscriminability and Topology}\label{subs:defs_egs}
The ability to decide whether one stimulus/input  is distinct from another is essential for adaptation, survival, and intelligent life. Intelligent agents are uniquely capable to activate knowledge to judge  distinction. Williamson calls this context-relative  process  discrimination (\cite{williamson1990identity}). Discrimination establishes a context-relative binary relation denoted by $\pind$ and called   \newterm{\adiakrisia}:\footnote{The \textgreek{ad} in $\pind$ stands for the first two letters of the Greek word for {\em indiscriminable} (\textalpha\textgreek{dia}\textkappa\textgreek{rit}\textomega\textvarsigma).} \footnote{
Context and its role in discrimination and similarity  judgments have been studied extensively and by many authors including \cite{tolman1932purposive,tversky1977features,vosniadou1989similarity,medin1993respects,raffman2000perceptual,gardenfors2004conceptual,
decock2011similarity,hebart2020revealing}.}

\begin{definition}[\cite{williamson1990identity}]\label{def:indis} Two inputs $x$ and $y$ are called \newterm{ \adiak}\ to a subject at a time $t$  if and only if at time $t$  the subject is not able to activate (acquire or employ) the relevant kind of knowledge that $x$ and $y$ are distinct.
\end{definition}
 Some authors refer to \adiakrisia\  as \newterm{active \adiakrisia}, see \cite{farkas2010subject}. The \adiakrisia\ relation is symmetric and reflexive and  generates a context-relative topology on the set of inputs $\X$. \footnote{Poincar\'e discusses \adiakrisia\ in \cite{poincare1930dernieres} but refers to it as {\em indiscernability}. Similarly Poston studies  {\em indistinguishibility}, basing  it on the ``limit of discrimination" of the biological senses and instruments, \cite{poston1971fuzzy}. }

\begin{definition} \label{def:doppelgangers+perceptual_topo} For every  $x\in \X$  let $\twins{x} = \left\{y\in \X \,  :\,   y\pind x\right\}$. A point  $y\in \twins{x}\setminus \{ x\}$  is called a \newterm{\dopp} [of $x$]. The \newterm{perceptual topology} $\taup$ is the topology generated by the sub-basis  $\mathfrak{D}_{\alpha\delta} = \left\{\twins{x} \right\}_{x\in \X}$. 
\end{definition}

\refstepcounter{e_counter}  \label{eg:optimal} {\bf Example  \ref{eg:optimal}} 
An input $x\in \X$ is called \newterm{optimal} if it does not have 
non-trivial/non-identical \dopps, i.e., if $\twins{x} = \{x\}$.\footnote{Optimal objects and light sources have been described and studied in  colorimetry, \cite{wyszecki2000color,Logvinenko2016Counting}.}  In particular, if $\pind$ is the identity relationship $=$ (i.e,. all inputs are optimal within the given context), then the perceptual topology $\taup$ is discrete. However, the  finiteness of  human  observations  (they are subject to finite time and finite work constraints) and the {\em bounded rationality} constraints imposed by the limitations on the availability of  information and computational capabilities to humans, \cite{simon1957models},  indicate that the scenario $\twins{x} = \{x\}$ for every $x\in \X$ may be highly unlikely. 

Often reflexive binary relations are defined and discussed as coverings of the underlying space. Indeed, every reflexive binary relation (RBR) $\approx$ on $\X$ defines a  covering ${\displaystyle \left\{\mathfrak{g}(x) = \{y: y\approx x\}\right\}_{x\in \X}}$, of $\X$ and vice versa one can define reflexive binary relationships through coverings of $\X$. See Appendix, Section~\ref{sect:tolerance}. In particular, the perceptual topology, $\taup$, is a tolerable topology, cf., Definition \ref{def:R_topo} in  Appendix, Section~\ref{sect:tolerance}.

\refstepcounter{e_counter}  \label{eg:boyle} {\bf Example  \ref{eg:boyle}:}
 In a series of experiments in 1875 William Thiery Preyer studied  human pitch discrimination,  
see \cite{preyer1876grenzen} and  also \cite{ellis1876sensitiveness}. They showed evidence that if $\X$ is the interval of frequencies  of pure sounds in an octave, then $\twins{x}$ is an interval whose diameter $l(x)$  depends on $x$. 
%In this work the volume of the sound was kept constant. 
\vskip 6pt

The active psychophysics research  on just noticeable difference initiated  by Weber and Fechner, \cite{weber1831pulsu,weber1846tastsinn,fechner1860elemente,fechner1882revision},  provides one of the few classes of examples where we have  empirically supported understanding of the perceptual topology. \\

\refstepcounter{e_counter}  \label{eg:Weber} \noindent {\bf Example  \ref{eg:Weber}:} Let $\X$ be the  closed bounded interval $[a, b]\subset (0, +\infty)$. Suppose that Weber's law holds  and let $k>0$ be the Weber constant. Let $w = 1+k$,  then 
\begin{equation} \label{eq:weber}
\twins{x} = \begin{cases}
  \left[a, xw\right), & a\leq x <  aw \\
   \left(x/w, xw\right), & aw\leq x \leq  b/w \\
     \left(x/w, b\right],  & b/w < x \leq  b. 
 \end{cases}
\end{equation}
The covering $\displaystyle\left\{ \twins{x} \right\}_{x\in [a, b] }$ defines a symmetric RBR but the relation is not transitive. The corresponding perceptual topology is $T_0$ but not $T_1$, and the topology is not pseudometric.

The transitivity or more more often the  lack of transitivity of $\pind$  have been studied extensively and proven or  postulated in many human experiences,  \cite{goodman1951structure,wright1975coherence,williamson1990identity,graff2001phenomenal,de2004perceptual,hellie2005noise,
pelling2008exactness,farkas2010subject,raffman2012indiscriminability,arstila2012transitivity,broderick2023foveated}.    

\begin{definition} \label{def:closure}  We will  denote by  $\sorites$ the  transitive closure of the \adiakrisia\ relation  $\pind$ on $\X$. It is defined explicitly as $x\sorites y$ iff  there exists a finite chain of \dopps\ $x=x_0 \pind  x_1\pind x_2\pind\cdots \pind x_{n}= y$.\footnote{The \textsigma\   in $\sorites$ is the first letter of the Greek word for {\em heap} (\textsigma\textomega\textgreek{r\'{o}}\textvarsigma).} We will call the relation $\sorites$ perceptual \newterm{metamorphy} and will refer to any two inputs $x \sorites y$ as \newterm{metamorphic}. 
Extending  Pawlak's terminology, \cite{pawlak1981classification} 
we call the equivalence classes in $\xmod$ (perceptually) \newterm{elementary sets}.
\end{definition} 

\refstepcounter{e_counter}   \label{eg:transit} \noindent {\bf Example  \ref{eg:transit}:} If the indiscriminability relation is transitive, then each $\twins{x}$ is an elementary set.  The perceptual topology may be optimal (recall Example \ref{eg:optimal}) or not. In the former case it is  Hausdorff and in fact $(\X,\taup)$ is a discrete manifold,  in the later case the topology $\taup$ is not $T_0$. In the former case 
 \begin{equation} \label{eq:pseudo}
d_{\cal K}(x,y) = \begin{cases}
  0, & x\pind y \\
  1, & x\not\pind y
 \end{cases}
\end{equation}
is a metric generating the  optimal perceptual topology, in the  later case $d_{\cal K}$ is a non-separating pseudo-metric generating the perceptual topology.\footnote{$d_{\cal K}(x, y) = d(x, y)({\twins{x})}$ where $d$ is the continuity defined by Kopperman in \cite{kopperman1988all}. }

Human visual perception provides a fundamental example where $\twins{x} \not= \{x\}$. In particular, if two images $x$ and $y$ differ only in unattended regions for example due to low saliency values (cf.  \cite{zhaoping2007psychophysical}), then $x\pind y$.  Visual metamers have been studied by many authors, including 
\cite{grassmann1853theorie,ostwald1919physikalische,wyszecki1953valenzmetrische,thornton1973matching,krantz1975color,cohen1985color,
balas2009summary,freeman2011metamers,logvinenko2013object,
Logvinenko2016Counting,feather2019metamers,jagadeesh2022texture,broderick2023foveated}.    In these studies, the input space  is assumed to be endowed with  Grassmann structure (see \cite{krantz1975color}), and in particular, the  indiscriminability relation is transitive.\footnote{In these studies, indiscriminable inputs are referred to as ``matching" or ``metameric", or ``alike".} 

\subsection{Indiscriminability, Indiscernibles,  and Discriminative Feature Representations}\label{subs:leibniz}
Let $\Phi$ be the space of all features of the inputs/stimuli $x\in \X$ and let $\Phi_x\subset \Phi$ be the set of features attributed to $x$ in a given context. The features attributed to $x$ are either  perceptually \newterm{robust}, i.e., features that belong to   $ \mathop{\bigcap}\limits_{y\in \twins{x}} \Phi_x\cap \Phi_y$, or perceptually \newterm{adversarial}, i.e., features that belong to $\mathop{\bigcup}\limits_{y\in \twins{x}} \Phi_x\setminus \Phi_y $.\footnote{Perceptually robust features are shared among all \dopps.  Intuitively robust features are features ``that are  robust on the noise (variation) and are invariant to the existence of the adversarial perturbation", \cite{kim2021distilling}. A feature is adversarial if it distinguishes $x$ from a \dopp.}  

\begin{definition}\label{def:idisc}Following \cite{forrest2020identity},  we say that $x$ and $y$ are \newterm{indiscernible}, in a given context, if  $\Phi_x = \Phi_y$.\footnote{Leibniz discussed indiscernability and postulated the Principle/Law of Identity of Indiscernibles, $(\Phi_x = \Phi_y)\implies (x = y)$, in  \cite{von1902leibniz} and in the third and fourth papers addressed to Samuel Clarke, \cite{clarke1717collection}.}  
\end{definition}

\begin{definition}\label{def:llp}
We will say that the feature representation  \newterm{satisfies the  Law of Indiscriminability} if  $\Phi_x = \Phi_y$ implies $x\pind y$. 
\end{definition}

\refstepcounter{e_counter}   \label{eg:transit_LLP} \noindent {\bf Example  \ref{eg:transit_LLP}:} Suppose that  the indiscriminability relation is transitive. Let $\Phi = \X$. The feature representation $\Phi_x = \twins{x}$ for every $x\in \X$, \LLP. Indeed, the transitivity of $\pind$ implies that $\twins{x} = \twins{y}$ iff $x\pind y$, and hence $\Phi_x=\Phi_y$ iff $x\pind y$.  

\begin{obs}\label{obs:law of indiscernibles} If there exists a feature representation that \LLP\ and such that $(x\pind y)\implies (\Phi_x = \Phi_y)$ , then $\pind$ is transitive.\footnote{See Appendix, Subsection \ref{subs:schreider} for the proof.} 
\end{obs}

The attempts to understand the role of feature representations in the processes of discrimination and, in particular, establishing identity are ongoing. Much remains unknown, reduced to embracing items of faith, and is frequently and thoroughly revised. See for example, \cite{black1952identity,perala2018aristotle}. 

One can hypothesize the existence of biologically-plausible perceptually \newterm{\dfrs}, i.e., context-relevant feature attributions $\Phi_{x}$, to the inputs $x\in \X$, such that 
\begin{equation}\label{eq:schreider}
\Phi_x \bigcap \Phi_y \neq \emptyset\Longleftrightarrow x\pind y.
\end{equation} 

For every discriminative feature representation, indiscernible inputs are indiscriminable, i.e., $(\Phi_x = \Phi_y)\implies (x\pind y)$. 
The feature representation in Example \ref{eg:transit_LLP} is a  discriminative feature representation and  \LLP. However, unless the corresponding perceptual topology is optimal, the Leibniz Law of Identity, $(\Phi_x = \Phi_y)\implies (x = y)$ does not hold. The Leibniz Law of Identity does hold in the following example.

\refstepcounter{e_counter}  \label{eg:Weber_R_plus} \noindent {\bf Example  \ref{eg:Weber_R_plus}:} Let $\X=(0, +\infty)$. Suppose that Weber's law holds  and let $k>0$ be the Weber constant. Let $w = 1+k$,  and let
\begin{equation} \label{eq:weber_R_plus}
\twins{x} = 
   \left(x/w, xw\right)
\end{equation}
The covering $\displaystyle\left\{ \twins{x} \right\}_{x\in\X }$ defines a symmetric RBR on $\X$. Let $\Phi$ be the collection of all sub-intervals in $\X$.  For every $x\in \X$, let $\Phi_{+}(x)\subset \Phi$ be the collection of all semi-closed intervals of the form  $[b, bw)$, $x/w < b\leq x$ and let $\Phi_{-}(x)\subset \Phi$ be the collection of all semi-closed intervals of the form  $(b/w, b]$, $x\leq b < xw$. Then, the assignment  $\Phi_x = \Phi_{-}(x)\bigcup\Phi_{+}(x)$ defines a discriminative feature representation on $\X$ where the  $\pind$ relationship is generated by the covering defined in Equation \ref{eq:weber_R_plus}. Furthermore, the Leibniz Law of Identity holds and the feature representation \LLP.

The existence of \dfrs\ is a specific instance of a general property of reflexive and symmetric binary relations.  
\begin{definition}\label{def:gdfr}
Let  $\approx$ be a reflexive and symmetric binary relation on a set $\X$, and let $\Phi$ be a set (of features). A feature representation 
$\{\Phi_x \subset \Phi\}_{x\in \X}$ is called a $\approx$-\newterm{\dfr}   if
\begin{equation}\label{eq:gen_dfr}
\Phi_x \bigcap \Phi_y \neq \emptyset\Longleftrightarrow x\approx y.
\end{equation}
\end{definition}
\begin{theo}\footnote{The  theorem is attributed to Kalmar and Yakubovich, and is proven for reflexive and symmetric binary relations on finite sets in \cite{schreider1975equality}. For a very simple proof  of the general case, that does not rely on transfinite induction,  see Subsection \ref{subs:schreider}  in the Appendix.}\label{th:dfr exits} 
Every symmetric and reflexive binary representation $\approx$ admits a   
$\approx$-\dfr. Specifically, there exists a set of features $\Phi$ and a  feature representation 
$\{\Phi_x \subset \Phi\}_{x\in \X}$ satisfying  Condition (\ref{eq:gen_dfr}). 
\end{theo} 
\Dfrs\ are just $\pind$-\dfrs. The attributed discriminative features\footnote{A feature $\xi\in \Phi$ is called \newterm{attributed} if $\xi\in \Phi_x$ for some input $x\in\X$. It is plausible that $\Phi = \mathop{\cup}\limits_{x\in \X} \Phi_x$, and so all features are attributed. However, many models do not preclude the existence of  spurious latent traits.} represent  structures of \dopps. Indeed, let $\{\Phi_x\subset \Phi\}_{x\in \X}$ be a a feature representation and let $\clu{\xi}$ be the context-dependent \newterm{semantic cluster} of inputs sharing the feature $\xi\in \Phi$. Specifically,  
\begin{equation}\label{eq:clx}
\clu{\xi}=\{x\in \X, \mbox{ s.t., } \xi\in\Phi_x\}.\footnote{The semantic cluster of inputs sharing the feature $\xi\in \Phi$ is defined for any feature representation.  A  feature $\xi$ is attributed in a given context, iff $\clu{\xi}\neq \emptyset$; a feature is a \newterm{hypothetical feature}, when  $\clu{\xi}= \emptyset$. See further discussion in Appendix Part \ref{ref}.}
\end{equation}

In particular, if  $\{\Phi_x\subset \Phi\}_{x\in \X}$ is a \dfr, then every attributed discriminative feature $\xi\in \Phi_x$ is associated to, and in some way explained  by, a collection of \dopps\ since $\clu{\xi}\subset \twins{x}$. 

\noindent {\bf Example  \ref{eg:transit_LLP} Continued:}    \label{eg:transit_LLP_clx} When indiscriminability is transitive, the feature representation $\{\Phi_x = \twins{x}\}_{x\in \X}$ is a \dfr, the \dopps\  $y\pind x$ of every  $x\in \X$ are the discriminative features of $x$  and 
\[
\clu{y} = \twins{x}, \forall y\in \twins{x}.
\]

The biological plausibility of discriminative feature representations is an open question.  Example \ref{eg:transit_LLP} and Example \ref{eg:Weber_R_plus} show that building discriminative feature representations might be too resource demanding, because, the size of the feature representations $\Phi_x$ may be too large. On the other hand,  even such large feature representations may be   starting points in the search and identification of smaller representations through a refinement process outlined in Appendix Part \ref{ref}. In particular, applying this process to the \dfr\  in Example \ref{eg:transit_LLP}, yields a new \dfr\ 
$\{\hat{\Phi}_x =\{ \twins{x}\}\}_{x\in \X}$. The new \dfr\ appears simpler, the feature representation of each input is a single feature. However, the new features are much more complex than the original features.  

\subsection{Contrast and Indiscriminability}\label{subs:contrast} 
Intuitively, indiscriminability is related to similarity or equivalently to contrast. Leibniz refers to indiscernibles as {\em perfectly alike} ({\em parfaitement semblables}),  \cite{clarke1717collection}, while  Poincar\`e, calls them {\em semblables \`el\`ements}, \cite{poincare1930dernieres}.  A common approach is to use geodesic distance to quantify similarity/contrast. However, empirical data shows that in  many experiences similarity/contrast are not symmetric or do not satisfy the triangle inequality. 

On the other hand, every  binary relation  is defined by a \newterm{contrast context}, that is a pair of  proset-valued maps 
 $(c, \epsilon)$. Indeed, let $\preccurlyeq$ be a pre-order on a set 
 ${\cal C}$, every pair of mappings  
 $c : \X\times \X \rightarrow {\cal C}$. and   
$\epsilon : \X\rightarrow {\cal C}$  defines a binary relation  $\picoco$ on $\X$ s.t. 
\begin{equation}\label{eq:equiv}
\picoco    =  \left\{ (x,y)\in \X\times\X : c(x,y)\preccurlyeq \epsilon(x) \right\}
\end{equation}
Indeed, if $\approx$ is a binary relation, then the contrast function  defined by 
\begin{equation} \label{eq:gen_cc}
c(x,y) = \begin{cases}
  0, & \mbox{ if } x\approx y \\
  1, & \mbox{ otherwise}, 
 \end{cases}
\end{equation}
and $\epsilon(x) = 1$ for all $x\in \X$ are $\{0,1\}$-valued and the pre-order relation is the usual $<$ do define the binary relation $\approx$. There are many contrast contexts that generate the same binary relation. 
A straightforward computation confirms that: 
\begin{obs}\label{obs:context corollary}
A binary relation is symmetric and  reflexive if and only if it is defined by a contrast context $(c, \epsilon)$ such that $c(x,x)\preccurlyeq \epsilon(x)$ for every $x\in \X$ and $(c(x,y)\preccurlyeq \epsilon(x) )\implies (c(y,x)\preccurlyeq \epsilon(y) )$ for every $x, y\in \X$. 

 A binary relation is an equivalence relation if and only if it is defined by contrast context $(c, \epsilon)$ such that $c(x,x)\preccurlyeq \epsilon(x)$ for every $x\in \X$,  $(c(x,y)\preccurlyeq \epsilon(x) )\implies (c(y,x)\preccurlyeq \epsilon(y) )$ for every $x, y\in \X$, and $(c(x,y)\preccurlyeq \epsilon(x) )\land((c(y,z)\preccurlyeq \epsilon(y) ) \implies (c(x,z)\preccurlyeq \epsilon(x) )$ for every $x, y, z\in \X$.
\end{obs}
%\begin{definition}\label{def:contrast} 
%Let $\preccurlyeq$ be a pre-order on a set ${\cal C}$, a pair of mappings   $\epsilon : \X\rightarrow ({\cal C}, \preccurlyeq)$  
%
%%\end{definition}
%\newterm{contrast evaluation context}
%\begin{obs}\label{obs:contrast} Every pair of mappings $(c, \epsilon)$  $\epsilon : \X\rightarrow ({\cal C}, \preccurlyeq)$ 
%\end{obs}
%\cite{poincare1930dernieres}
%

\refstepcounter{e_counter}   \label{eg:distance_tolerance} \noindent {\bf Example  \ref{eg:distance_tolerance}:}  The metric tolerance relation introduced by Zeeman (see \cite{Zeeman1965Topology,zeeman2017tolerance})  is defined by the contrast context $(c(x,y)=\mbox{dist}(x, y), \epsilon(x) = \epsilon)$, where $\mbox{dist}$ is a distance on $\X$, $\epsilon>0$ is a fixed real constant and once again $({\cal C}, \preccurlyeq)$ is $([0,\infty), <)$. 

\refstepcounter{e_counter}   \label{eg:pawlak} \noindent {\bf Example  \ref{eg:pawlak}:} Pawlak's framework for object classification ``by means of attributes", \cite{pawlak1981classification}, describes an approach to indiscriminability.  The key ingredients of the framework are   collection $A$ of attributes, i.e.,   maps $a: \X\rightarrow \frak{F}_a$ assigning attribute values $a(x)$ to inputs $x\in \X$ and a  knowledge function $\rho: \X\times A\rightarrow \displaystyle{\prod_{a\in A}} \frak{F}_a$, such that $\rho(x,a)\in \frak{F}_a$ for all $x\in \X$. The product of all attribute value ranges is the collection of all possible features attributable to the inputs cf. Section \ref{subs:leibniz}, that is,  
$\displaystyle{\prod_{a\in A}} \frak{F}_a   = \Phi$, and the $a$-th component of $\Phi_x$ is  $\rho(x,a)$. 

The indiscriminability relation defined by Pawlak is just the transitive indiscernibility relation $x\pind y$ iff $\Phi_x=\Phi_y$ and is generated by the contrast context $(c, \epsilon)$ where 
\begin{equation} \label{eq:pawlak}
c(x,y) = \begin{cases}
  0, & \mbox{ if } \Phi_x = \Phi_ y \\
  1, & \mbox{ if } \Phi_x \neq \Phi_ y 
 \end{cases}
\end{equation}
and $\epsilon(x) = 1$ for all $x\in \X$ are $\{0,1\}$-valued and the pre-order relation is the usual $<$. 

\cite{peters2007near,peters2012applications,peters2012tolerance} present a historic overview of  near sets, tolerance spaces, rough sets, indiscernability, and  indistinguishibility. In particular, Peters and his students propose that two inputs are indiscrminable if and only if their features are sufficiently similar where similarity is measured by a metric on a (finite dimensional)  space of features (probe measurements) $\Phi$ (see \cite{peters2012tolerance}). Namely, $x\pind y$ iff 
$\mbox{dist}(\Phi_x,\Phi_y) < \epsilon$, where  $\epsilon >0$ is a fixed context-dependent tolerance level. Generically, this  indiscriminability  relation is not transitive. It is  defined by the contrast context $(c(x,y) = \mbox{dist}(\Phi_x,\Phi_y), \epsilon(x)\equiv \epsilon)$.  

Contrast context reflects our common experience that we judge and discriminate things and ideas based on the differences between them, not on the distance between them. It provides a general approach to the study of indiscriminability. For example, the distance tolerance studied by Zeeman and Peters (Example \ref{eg:pawlak}) cannot be used to obtain the perceptual topology generated by Weber's law. However, defining the proper contrast context captures the subtleties of Weber’s law, and defines the relevant perceptual topology.

\refstepcounter{e_counter}   \label{eg:weber_contrast} \noindent {\bf Example  \ref{eg:weber_contrast}:} 
Let $\X$ be the  closed bounded interval $[a, b]\subset (0, +\infty)$. Suppose that Weber's law holds  and let $k>0$ be the Weber constant. The \adiakrisia\ relation described in Example \ref{eg:Weber} is defined by the  contrast context $(c, \epsilon)$ where: 
\begin{equation} \label{eq:weber_contrast}
c(x,y) = \begin{cases}
  (y-x)/x, & x\leq y \\
 (x-y)/y, & x\geq y  
 \end{cases}
\end{equation}
and $\epsilon(x) = k$ for all $x\in \X$ are $[0,\infty)$-valued and the pre-order relation is the usual $<$.

\section{Classifiers, Adversarial \dopps.} \label{sec:Class}

In this paper we identify classifiers with finite segmentations of the space of inputs $\X$ since the labeling function $\mbox{label}_R : \X\rightarrow \{1,\ldots,m\}$ of a classifier $R$ partitions $\X$ into disjoint label sets 
\[
R_c = \left\{x\in X : \mathop{\mbox{label}_{R}}(x)=c\right\},\hspace{1em} c=1,\ldots,m
\]
and
\begin{equation}\label{eq:class func}
\labelR{x} =\sum_{c=1}^{m} c\charf{R_{c}}(x), \forall x\in\X.
\end{equation}
The segmentation $R$ is called \newterm{fully populated} iff the labeling function $\mbox{label}_R$ is surjective mapping onto the range of labels $\{1,\ldots,m\}$. 
\subsection{Adversarial \dopps} \label{subs:ad}
\begin{definition}\label{def:ad_dopp} We say that $x$ is a \dopp\ adversarial to the classifier $R$ iff  $\exists y \in \twins{x}$ such that $\mbox{label}_R(x)\neq \mbox{label}_R(y)$ and we will refer to both $x$ and $y$ as \newterm{adversarial \dopps} when the classifier $R$ is clear from the context. 

A classifier $R$ is called \newterm{(perceptually) regular} iff it does not admit adversarial \dopps.  

We say that the \newterm{classification problem with $m$-labels is well defined} if there exists a fully populated and perceptually regular classifier $R=\{R_1,\ldots, R_m\}$ with $m$ labels. Otherwise we say that the classification problem with $m$-labels is \newterm{not well defined}.

If $R$ is regular, then $\mathfrak{D}_{\textgreek{ad}} = \left\{\twins{x} \right\}_{x\in \X}$ are  $R$ coherent coverings (as defined in \cite{schreider1975equality}). However, coherency does not imply regularity. Namely if $\mathfrak{D}_{\textgreek{ad}}$   is  coarser than the covering defined by $R$, then $\mathfrak{D}_{\textgreek{ad}}$  and $R$ are coherent but the later may admit \dopps.   
\end{definition}

The labeling function $\mbox{label}_R$ of a regular classifier is  continuous with respect to the perceptual topology $\taup$, i.e., $\mbox{label}_R\in \mathop{C^0}(\X,\taup)$ and harmonic with respect to a specific perceptually-based  Laplace operator $\slap$ defined in Section \ref{sec:distance}. Furthermore, if $R$ is not regular and $x$ is a point of discontinuity of $\mbox{label}_R: (\X, \taup)\rightarrow \{1,\dots,m\}$ then $x$ is an adversarial \dopp, 
i.e., there exists $y\pind x$ such that $\labelR{x}\neq\labelR{y}$. However, $y\pind x$ such that $\labelR{x}\neq\labelR{y}$ does not imply that either $x$ or $y$ is a point of discontinuity of $\mbox{label}_R$.  On the other hand, if every two points $x \pind y$ are not separable,\footnote{For example, if $\pind$ is transitive as postulated in \cite{broderick2023foveated}} then every \dopp\ adversarial to $R$ is a discontinuity of $\mbox{\rm label}_R$ or equivalently $R$ is regular if and only if $\mbox{\rm label}_R\in \mathop{C^0}(\X,\taup)$.

It is well known that in some  experiences perceptually unambiguous categories and hence perceptually regular classifiers do  not exist. A simple  example is provided by the perceptual topology  in Example \ref{eg:Weber}. Indeed, in this case $\xmod$ is a singleton, i.e, every two inputs are metamorphic and hence there is a only one perceptually elementary set which equals the whole  $\X$.   Therefore,   every classifier with two or more labels must have adversarial \dopps.\footnote{See Lemma \ref{lemma:web} and the short argument that follows it in Appendix, Section \ref{sect:sorites}.} On the other hand, the following example shows that regular models do exist: 

\refstepcounter{e_counter}   \label{eg:Weber2} \noindent {\bf Example  \ref{eg:Weber2}:}  Let  $w > 1$ and $\X$ be the  closed bounded interval $[a, b']\subset (0, +\infty)$ and let $a<b<b'/w$    and 
\begin{equation} \label{eq:weber2}
\twins{x} = \begin{cases}
  \left[a, xw\right), & a\leq x <  aw \\
  \left(x/w, xw\right), & aw\leq x \leq b/w \\
  \left(x/w, b\right],  & b/w < x \leq  b \\
   \left(b, xw\right), & b < x\leq  bw \\
  \left(x/w, b'\right],  & bw < x \leq  b'. 
      \end{cases}
\end{equation}
There exists a unique regular fully populated  classifier with two labels. Every other fully populated classifier with two or more labels must admit adversarial \dopps.
\vskip 12pt

Clearly, no amount of ``robust training" will get rid of adversarial examples of a classifier with a surjective labeling function  $\labelf{R} : (\X, \taup) \rightarrow\{1,\ldots,m\}$ if $\X$ cannot be broken into $m$ perceptually unambiguous categories. In the rest of this section we investigate the non existence, existence and internal structure of regular classifiers.

\myeg{trans} If $\pind$ is transitive then for every number of labels $m$ smaller than the number of equivalence classes $\card{\X/\!\!\pind}$ there exists a fully populated regular classifier with $m$ labels, but if the number of labels $m$  is bigger than  $\card{\X/\!\!\pind}$ then every fully populated classifier with $m$ labels must admit \adv\ \dopps. 
\vskip12pt

Example \ref{eg:trans} indicates   that  if the transitive closure $\sorites$ is trivial,i.e., $\sorites = \X\times\X$, then no label is safe. Namely: 

\begin{obs}\label{claim:not_safe}
If  the transitive closure $\sorites$  of the \adiakrisia\ relation $\pind$  is trivial, then every fully populated classifier with two or more classes admits \adv\ \dopps. In particular, let $R$ be a fully populated classifier  with a surjective  labeling function $\mbox{label}_R : X\rightarrow \{1, 2, \ldots,m\}$, then for every label $c$ there exist \adv\ \dopps\  $x(c)\in \X$ and $ x^{*}(c)\in \twins{x(c)}$  such that $c=\labelR{x(c)}$ and $\labelR{x^{*}(c)}\neq\labelR{x(c)}$. 
\end{obs}

The proof follows from the fact that every finite chain of \dopps\  connecting points that are labeled differently by a classifier must contain a pair of adversarial  \dopps. See Lemma \ref{lemma:chain}  in Appendix Section \ref{sect:sorites}.
%This fact is related to the  {\it instability hypothesis}, \cite{raffman2012indiscriminability}. 

\subsection{Regular models: Existence and structure}

A straight-forward argument shows that  $R = \{R_1,\ldots, R_m\}$  is a  perceptually regular fully populated  classifier, then $x\in R_i$ iff $ [x]_{\sorites} \subset R_i$, i.e., each segment is a disjoint union of equivalence classes $ [x]_{\sorites}$ and so using the pigeonhole  principle  we obtain:
\begin{obs}\label{claim_pigeon_hole}
If the number of equivalence classes   $\card{\xmod}\geq 2$, then for every natural number  $2\leq m\leq \card{\xmod}$ there exists a perceptually regular fully populated classifier with $m$ labels. However,  if $m >  \card{\xmod}$, then every fully populated classifier with $m$ labels must have \adv\ \dopps. 

In particular, if $p= \card{\xmod}\geq 2$ is finite, then for every natural number $m\leq p$ there are exactly $S(p,m)$ regular fully populated classifiers with $m$ classes.  Here $S(p,m)$ is the Sterling number of the second kind.
\end{obs}

Observation \ref{claim_pigeon_hole} shows that the problem of finding a fully populated perceptually unambiguous classifier   $R = \{R_1,\ldots, R_m\}$ with precisely $m$ labels  is not well defined if $\card{\xmod} <  m$ and vice versa that the same problem is well defined  for every $m$ such that $\card{\xmod}\geq m$. In the former case solutions do not exist while in the later case solutions exist and   each class segment $R_i$ is a union of equivalence classes $[x]_{\sorites}$, 
\begin{equation}\label{eq:regular cluster segment}
R_{i} = \bigcup_{x\in R_{i}} [x]_{\sorites}.
\end{equation}

The existence and properties of \dfrs\ provide  insight whether a classification problem is well defined. In particular,   if there exists a \dfr\    and the set of attributed features is finite, then every classification problem,   whose number of labels exceeds the number of attributed features, is not well defined. See Observation \ref{ob: finite dfr}
in Appendix Part \ref{subs:app:finite}. 

Each label category $R_c = \left\{x\in X : \mathop{\mbox{label}_{R}}(x)=c\right\},\hspace{1em} c=1,\ldots,m$, where $R$  is a regular classifier, can be thought of as a unambiguous perceptual or semantic category. Various attributes including structural entropy, the overall similarity of elements within a category, and further structural components as the category core and fringe are discussed in Section \ref{sec:similarity}.

%\subsection{Misclassification: Failure to generalize and perceptual failure}
%Failure to generalize: $\labelR{x}\neq \labelR{y}$ and $x\not \sorites y$. 
%
%\noindent Perceptual failure: $\labelR{x}\neq \labelR{y}$ and $x \sorites y$.
\section{ Accuracy  and Adversarial \dopps. }\label{section:accuracy} 
The accuracy-adversarial robustness trade off observed and discussed in the literature involves various measures of accuracy \cite{su2018robustness,tsipras2018robustness,zhang2019theoretically,nie2022diffusion,omar2022robust}. We will discuss classification accuracy. We will show that there is a strong relationship between classifier  accuracy and vulnerability to adversarial \dopps. In particular, we will identify (perceptual) scenarios in which low accuracy  classifiers are critically vulnerable to adversarial \dopp\ attacks but on the other hand all high accuracy  classifiers can be fooled only by \dopps.   
\subsection{The probabilistic setup: recall rates, accuracy, bounds}
We will assume that $(\X, \F, \mu) $ is  a probability measure space equipped  with perceptual topology  $\taup$ generated by an indsicriminability relation $\pind$ such that for every $x\in \X$ the set of \dopps\   $\twins{x}$ and the equivalence class  $[x]_{\sorites} $ are events, ($\twins{x}\in \F$ and $ [x]_{\sorites}\in \F$, $\forall x\in\X$) .  

In the rest of this section we will assume that the classification problem with $m\geq 2$ labels is well defined and let  $\Omega = \{\Omega_1,\ldots, \Omega_m\}$  be a perceptually regular world model; we will reserve the notation $R = \{R_1,\ldots, R_m\}$  to denote any classifier ($R$ may or may not be perceptually regular) such that $R_i \cap \Omega_i$, $i=1,\ldots,m$ are the true positives (of class $i$). To define accuracy we will focus only on regular models and classifiers s.t.,  $\Omega_i\in \F$ and  $R_i\in \F$, for all $i=1,\ldots,m$. The \newterm{accuracy} of the classifier $R$ defined as 
\begin{equation}\label{eq:accuracy}
\mbox{accuracy}_\Omega(R) = \mu(R_1\cap \Omega_1)+\cdots +  \mu(R_m \cap \Omega_m).
\end{equation} 
Furthermore let us assume that  $\mu(\Omega_i) > 0$ for every $i=1,\ldots, m$ and thus we can define recall rates 
\begin{equation}
\rho_i = \frac{\mu(R_i\cap \Omega_i)}{\mu(\Omega_i)}, \hspace{1em} i=1,\ldots,m
\end{equation}
Bounds on the recall rates  imply bounds on the accuracy. Namely if 
\[
\underline{\rho} \leq  \rho_i \leq \bar{\rho}, \hspace{1em} i=1,\ldots,m
\]
then since $\mu$ is a probability measure on $X$, and 
\begin{equation}
\mu(R_i\cap \Omega_i) = \rho_i \mu(\Omega_i), \hspace{1em} i=1,\ldots,m
\end{equation}
we get  
\[
\underline{\rho} \leq  \mbox{accuracy}_\Omega(R)=\sum_{i=1}^{m} \rho_i \mu(\Omega_i) \leq \bar{\rho}.
\]
\subsection{\dopps,  misclassification\ldots but can we trade?} \label{subs:can}
Let $i(x)\in \{1,\ldots, m\}$  be the \newterm{object class label} of  $x\in \X$ i.e., $x \in \Omega_{i(x)}$ and 
\begin{equation}\label{eq:kbar}
\bar{k}(\Omega) = \sup_{x\in \X}\left(\frac{ \mu\left(\Omega_{{i(x)}}\right)}{\mu(\twins{x})}\right) .
\end{equation}
Every classifier whose recall rates do not exceed $1/\bar{k}(\Omega)$  is totally unsafe in the sense that every correctly classified input admits adversarial \dopps. Specifically: 
\begin{obs} \label{claim:no label is safe} Suppose that  the sets of \dopps\ are not negligible and   $\displaystyle{\inf_{x\in\X} \mu(\twins{x}) > 0}$, and let  $\classif{\Omega}{m}$ be a regular world model and let 
 $R = \{R_1,\ldots, R_m\}$ be a classifier whose recall rates are strictly  smaller than $1/\bar{k}(\Omega)$ and so 
 \begin{equation}\label{eq:low} 
\frac{ \mu\left(R_{i(x)}\cap\twins{x}\right)}{\mu(\twins{x})} \leq   \bar{\rho}\bar{k}(\Omega)<1.
\end{equation}
 Thus every  correctly classified input $x$ has adversarial \dopps. 
\end{obs}
Observation \ref{claim:no label is safe} shows that sacrificing accuracy may  lead to increasing the probability of encountering adversarial \dopps\ and in fact that there is no trade off for accuracies that are sufficiently low provided that all sets of \dopps\  have positive measure.  

The lack of an opportunity for a trade-off is even more striking when one tries to improve high recall rate (and hence high accuracy) classifiers. 
\begin{obs}\footnote{The proof is in Appendix Part\ref{sect:adv}.} \label{claim:miss_class_adv_dop} If  $\displaystyle{\inf_{x\in\X} \mu(\twins{x}) > 0}$, and let 
 $R = \{R_1,\ldots, R_m\}$ be a classifier whose recall rates are sufficiently high so that  $\underline{\rho}>1-1/\bar{k}(\Omega)$. i.e.,
\begin{equation}\label{eq:high}
\left(1-\underline{\rho}\right)\bar{k}(\Omega)<1.
\end{equation}
Then every   misclassified input $x$ is an adversarial \dopp. 
\end{obs}
\begin{definition}\label{def:humanlike} We say that a classifier has a \newterm{hyper-sensitive behavior} if every misclassified input is an adversarial \dopp.
\end{definition}
For classifiers with hyper-sensitive behavior  adversarial  robustness can only be improved by improving accuracies, i.e., by eliminating misclassification. Observation \ref{claim:miss_class_adv_dop} shows that if $\mu(\twins{x}) > 0$, for all inputs $x\in \X$,  then all classifiers with sufficiently high accuracy   are either regular (when accuracy equals to one) or have hyper-sensitive behavior. 
  
In summary  adversarial \dopp\  robustness - accuracy trade-off may happen for classifiers with middling accuracy rates or when  there are inputs whose  \dopps\  are negligible in measure. 
\section{Are the differences between  \dopps\ small?} \label{sec:distance}

\refstepcounter{e_counter}   \label{eg:inisible} \noindent {\bf Example  \ref{eg:inisible}:} 
Well established phenomena including the  bias  toward  same response in perceptual matching ( \cite{cunningham1982visual, krueger1978theory}),  the own-race bias ( \cite{meissner2001thirty,tanaka2004holistic,goldinger2009deficits,tanaka2009neural}), and own-age bias ( \cite{hills2011own}) provide evidence that $\twins{x}$  may  not be small $\ell_p$  balls. 

Furthermore, there is a mounting evidence that  humans may perceive as visible light and even identify  the hue of light waves  in the so called invisible ranges of the spectrum. Near IR wavelengths  in the  $950 \mbox{nm}$ to  $1040 \mbox{nm}$  range are perceived as shade of red and that near IR wavelengths just above   $1040 \mbox{nm}$ are perceived as ``light with colors corresponding to roughly half of the excitation wavelengths",  \cite{palczewska2014human}. On the other hand,  \cite{hammond2018individual} reports  that some groups of young adults detect  $315 \mbox{nm}$ UV radiation and perceive it as a shade of violet. These studies provide an example of a  perceptual topology on $\X = [0,+\infty)$ whose sub-basis  includes neighborhoods  $\twins{x}$ which are not  connected with respect to the Euclidean topology on the ray  of wavelengths $\X$. 

It turns out that if $x\pind y$, then $y$ is a small perturbation of $x$ and vice versa in the sense that the distance between $x$ and $y$ is relatively small with respect to an appropriate metric  $\pdist{\cdot}{\cdot}$ on $\X$. The metric is built using the discrimination graph $\pgraph$ described in the next paragraph. The Laplace operator $\plap$ on $\pgraph$ may be somewhat quirky, for example, piecewise-constant functions may not be harmonic. At the end of this section  we discuss a better-behaved  context-relative Laplace operator $\slap$. The  salience measure introduced in \cite{tversky1977features} is harmonic with respect to $\triangle_{\sigma}$, (see Section \ref{sec:similarity}).

The indiscriminability relation $\pind$ defines a un-oriented graph $\pgraph$ where $\X$ is the set of vertices and we say that there is an edge $\{x, y\}\in E_{\alpha\delta}$  between the vertices $x, y\in \X$ iff $x\pind y$. We will call $\pgraph$ the \newterm{discrimination graph}. 
 Let $\epdist{x}{y}$ be the graph distance between the vertices $x$ and $y$ and, in particular,   $\epdist{x}{y}=\infty$ iff $x\not\sorites y$.\footnote{The existence of the extended metric was hinted at in \cite{meinong1896bedeutung}; it was  discussed in a  related  context in  \cite{russell2020principles} and  rediscovered and exploited in  \cite{poston1971fuzzy}.} 
 Thus 
\begin{equation}\label{eq:pdist}
\pdist{x}{y} = \frac{\epdist{x}{y}}{1+ \epdist{x}{y}}, \forall x,y\in\X
\end{equation}
defines a distance  $\pdistNA : \X\times \X\rightarrow [0, 1]$. 

\begin{definition}\label{def:pd}
We will call $\epdistNA$ the \newterm{extended perceptual distance} and $\pdistNA$ the \newterm{perceptual distance}. 
\end{definition}
The input space $\X$ into, at most countably many, disjoint spheres 
\[
S^{w}_{\rho} (x) = \left\{y\in X, \mbox{s.t.} \pdistNA(x,y) = \rho\in [0, 1]\right\}. 
\]

The open ball of radius one $ \mathring{B}^{w}_{1}(x) =  \left\{y\in X, \mbox{s.t.} \pdistNA(x,y) < 1\right\}$  is just the equivalence class $[x]_{\sorites}$ with respect to the equivalence relation $\sorites$. 
Some of the strata may be empty sets. For example, if $\sorites$ is trivial,  then $S^{w}_{1} (x) =  \emptyset$ and  $\X = [x]_{\sorites} =  \mathring{B}^{w}_{1}(x)= S^{w}_{0} (x) \bigcup S^{w}_{1/2} (x)$  for every $x\in \X$.  On the other hand, if $\pind$ is transitive ($\pind$ equals $\sorites$), then $[x]_{\sorites} =  \mathring{B}^{w}_{1}(x) = B^{w}_{1/2}(x)$ and $\X = S^{w}_{0} (x) \bigcup S^{w}_{1/2} (x) \bigcup S^{w}_{1} (x)$. The former case holds whenever the graph distance ($\epdistNA$)  between any two inputs is finite (three, six or whatever). An example of the later case is provided in \cite{broderick2023foveated}. 

More interestingly the \dopps\ of a point are precisely the nearest points to $x$ if the distance is measured by $\pdistNA$ or $\epdistNA$. 

\begin{figure}[h]  \label{fig:strata}
\begin{center}
%%\framebox[4.0in]{$\;$}
%\fbox{\rule[-.5cm]{0cm}{4cm} \rule[-.5cm]{4cm}{0cm}}
\includegraphics[height =5cm]{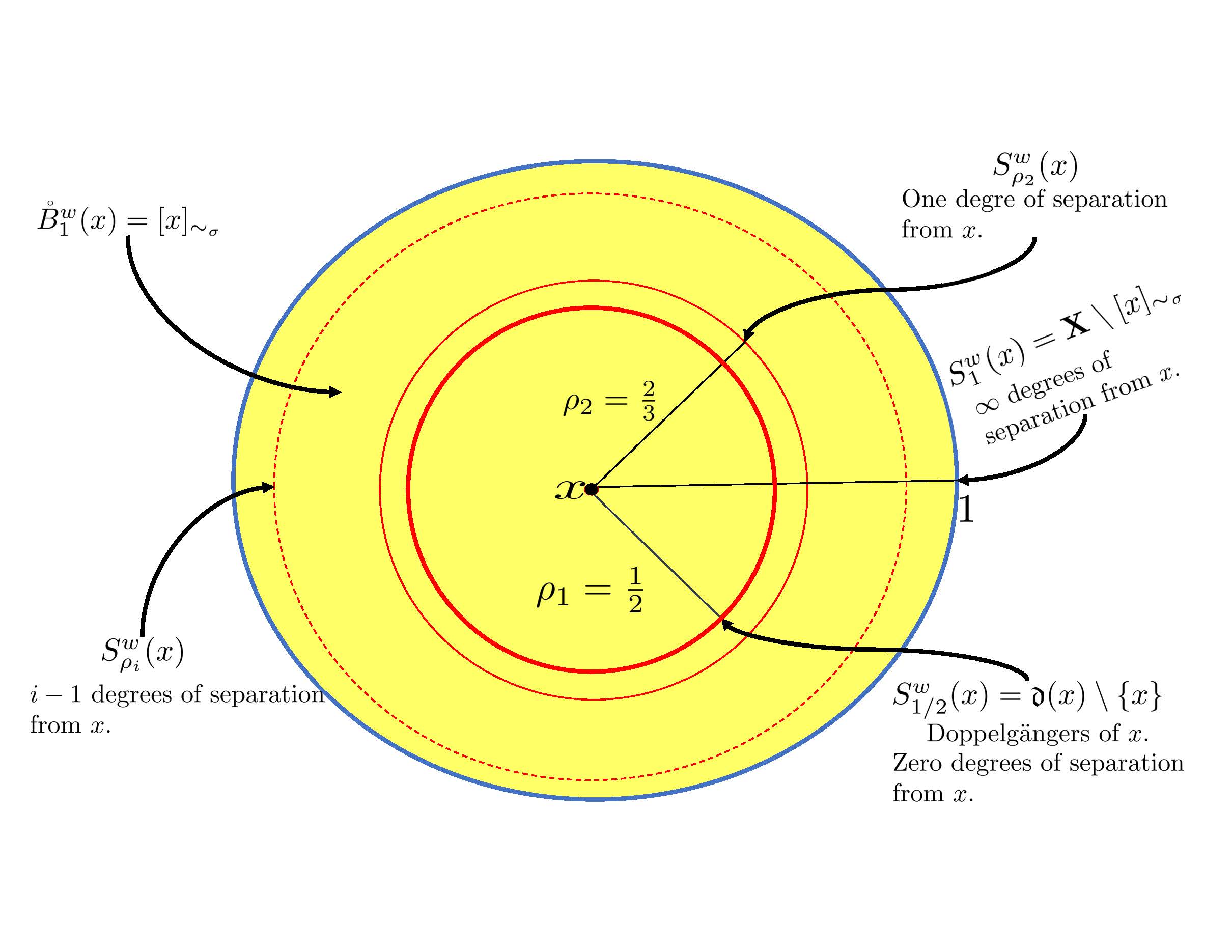}
\end{center}
\caption{Golyadkin's torment: ``From the view point"  of an input/stimulus  $x\in \X$, the space $\X$ is stratified  into concentric spheres, the nearest neighbors of $x$ are precisely its \dopps\ some or all of which may be adversarial.}
\end{figure}

It is easy to show that  in  Examples \ref{eg:Weber} and \ref{eg:Weber2}, where the probability measure $\mu$ is the uniform measure all nontrivial globally constant functions are not $\plap$ harmonic\footnote{For more detail see Section \ref{sect:laplace} in the Appendix.}. The  labeling functions of regular classifiers are step functions and belong to the class of perceptually regular functions,  i.e., functions that respect the regularity on the space of inputs imposed by the humans' inability to discriminate different "raw" signals.  
\begin{definition}\label{def:prl} A function $f: \X\rightarrow \R$ is called  \newterm{perceptually regular} if $f = \mbox{const}$ on every equivalence class  $\zeta\in \xmod$.\footnote{Thus $f$ is perceptually regular if and only if $f:(\X, \pind)\rightarrow (\R, =)$  is a morphism of tolerance spaces, \cite{sossinsky1986tolerance}, mapping the  (perceptual) tolerance space $(\X, \pind)$ into the optimal  tolerance space $(\R, =)$.}  Let $\displaystyle{L^1_{\mbox{pr}}(\X, \mu, \pind)}$ denote the vector space of perceptually regular integrable functions. 
\end{definition}

\myeg{plap_trans} If $\pind$ is transitive (as in \cite{broderick2023foveated} for example) and hence the graph $\pgraph$ is regular,  then $\ker{\plap} = \displaystyle{L^1_{\mbox{pr}}(\X, \mu, \pind)}$ and the spectrum of the Laplace operator is the set $\{0, 1\}$. 

The operator $\plap$ has at least three aspects that make it hard to use. First, it may not be well defined in many cases. Second, piecewise constant functions are not necessarily  $\plap$-harmonic. Third, it is hard to describe the full spectrum of $\plap$ for most perceptual topologies $\taup$. We will define another operator that exists in many situations when $\plap$ is not well defined. Furthermore, all perceptually regular functions are harmonic with respect to it. Finally, its spectrum is easy to compute. 
Assuming that the function $\sdeg : \X\rightarrow [0,+\infty)$ defined by 
\begin{equation}\label{eq:snode_degree}
\sdeg(x) = \mu([x]_{\sorites})
\end{equation}
 is positive, then we define the \newterm{\dopps\ chain Laplace operator} by:
\begin{equation}\label{eq:slap}
\slap(f) (x)= f(x) - \frac{1}{\sdeg(x)}\int_{[x]_{\sorites}}f
\end{equation}
Note that $\slap$ is defined whenever $\plap$ is defined and sometimes $\plap = \slap$. The spectrum of $\slap$ is the set $\{0, 1\}$,  
 \begin{equation}\label{wq:ker_slap}
 \ker{\slap} = \displaystyle{L^1_{\mbox{pr}}(\X, \mu, \pind)}
 \end{equation} 
and 
\begin{equation}\label{eq:slap ones}
 \ker(\slap - \mbox{Id}) = \left\{f   :  \int_{[x]_{\sorites}}f = 0, \forall x\in \X\right\} \neq \{0\}.
\end{equation}

%\noindent{\bf Note} Each equivalence class $[x]_{\sorites}$ is a maximal clique in the perceptually-based graph $\pgraph$ and in the perceptually-based graph $\sgraph$. 
%
%\begin{definition}\label{def:pclique}We will call the equivalence classes $\alpha\in \xmod$ \newterm{perceptual cliques}.
%\end{definition}
\section{Life without borders, similarity, core and fringe.}\label{sec:similarity}
%{Similarity, salience, contrast, prototypes,  and degrees of separation}
A important property  of the perceptual topology is that the if a classifier $\Omega$ is perceptually regular, then it ``imposes an open [topological] borders policy", that is,  $\partial \Omega_c = \emptyset$ for every label $c$. Linguists and psychologists, have observed and postulated that  natural perceptual and semantic categories are borderless. See for example  \cite{rosch1973internal}. Class/decision boundaries are studied and exploited  in many works on classifiers, and in particular on  adversarial robustness. These boundaries are artifacts of the metric topology used by the researchers, they are not perceptual phenomena.

However, we are all familiar with the idea that some stimuli are more intrinsic/representative to/of a given class and at the same time frequently there are objects/stimuli/inputs that, while they are firmly with in the class, are less representative/share few(er) features compared to  the rest of the elements in the class, that is, they "are/belong to  the fringe" of the class. 

\subsection{Affinity, core, and fringe} \label{subs:proto}

\begin{definition}\label{def:similarity}
Let $s$ be a similarity scale, i.e., a function $s:\X\times\X \rightarrow \R$ such that $s(x,x)\geq s(x, y), \forall x, y\in \X$ as in \cite{tversky1977features}, \cite{medin1993respects} (measuring similarity within a fixed context) and \cite{lin1998information}.
\end{definition}
The values $s(x,x)$ can be and sometimes are used to represent the \newterm{salience} or equivalently \newterm{importance} of the input within $\X$, see for example \cite{tversky1977features}. The similarity scale provides a method to quantify the  affinity of a input/stimulus to a given measurable subset $D\subset \X$ and the notions of prototype and fringe. 
%Similarity and prototypes play important roles classification, word representations, and the development and understanding of concepts.  
The ($s$-)\newterm{affinity}  of $x$ with a measurable set $D$ is defined as 
\begin{equation}\label{eq:total similarity}
\proto{x}{D} = \int_{D} s(x,y)
\end{equation}
This is a straightforward generalization of the notion of prtototypicality defined in \cite{tversky1977features}.
\begin{definition}\label{def:prototype} $x\in D$ is called a prototype of $D$ (with respect to the integrable similarity scale $s: \X \times \X \rightarrow \R$) if 
\begin{equation}\label{eq:proto}
\proto{x}{D} = \sup_{z\in D} \proto{z}{D}
\end{equation}

$x\in D$ is called a fringe element of $D$ (with respect to the integrable similarity scale $s: \X \times \X \rightarrow \R$) if 
\begin{equation}\label{eq:fringe}
\proto{x}{D} = \inf_{z\in D} \proto{z}{D}
\end{equation}
\end{definition}

One is tempted to think of prototypes as the stimuli that are ``clearest cases, best examples". ``easy to tell apart" and to "be a good representative" and hence  that optimize salience \cite{heider1971focal,rosch1973internal,rosch1975cognitive,
tversky1977features,douven2019putting}. However, clearly there is no reason to expect that a stimulus that is unlikely to be observed/encountered would be  selected as a prototype. The examples below show that prototypes and fringes are obtained by optimizing a mixture of the frequency of appearance (likelihood to encounter) and salience.  

The core and fringe sets may be empty. In practice one is often satisfied with finding elements that may not be true prototypes but belong to the \newterm{$M$-core}, i.e., have affinity exceeding  a fixed threshold $M$. Similarly, elements  whose affinity is below a given threshold $\tau$ belong to the  \newterm{$\tau$-fringe}.
\subsection{Features, salience, similarity, and class attributes and structure.}\label{subs:features}

Many perceptual and cognitive processes  exploit the intelligent agents' abilities to measure and/or compare the  salience/importance of features. In some simple cases it is expected and it even might be true  that salience/importance is a probability measure $f_\Phi$ on on the space of (all possible) features $\Phi$ and the feature representations $\Phi_x$ are measurable subsets of $\Phi$. In many accounts including Tversky's feature contrast model \cite{tversky1977features} the \newterm{salience scale} $f_\Phi$ is a context dependent,  nonnegative function defined on a collection $\Upsilon(\Phi)$ of subsets of $\Phi$ which is closed under finite unions and  intersections, and set differences. Furthermore,  $
\Phi_x\in \Ups, \forall x\in\X 
$; and the non-negative function $f_\Phi$ is 
\newterm{feature additive}, i.e, 
$
f_\Phi(A\cup B) = f_\Phi(A) +  f_\Phi(B), \mbox{if $A\cap B=\emptyset$}$. 
The value $
f(x) = f_\phi(\Phi_x), \forall x\in \X $ 
is called the \newterm{salience/prominence of the input} $x$, \cite{tversky1977features}. 
%\footnote{A possible mechanism for computing stimuli/input importance could be based on quantifying the activation levels of brain networks as the medial frontoparietal  network;   the face network involving regions in the frontal, temporal, and occipital cortex; the salience network rooted in the anterior insula. For example, this mechanism could exploit the subregions coherence matrix.} 
%whose functions in cognition were discussed in \cite{seeley2007dissociable,menon2010saliency,chang2013decoding}     may be a mechanism that supports the computation of features importance/salience.} 
We will call a  (Tversky) salience scale $f_\Phi$ perceptually regular if the prominence/salience function $f(x) =  f_\phi(\Phi_x)$ is perceptually regular. We will call a perceptually regular salience scale $f_\Phi$ \newterm{fully deployable} if it can be used to judge the distinguishing features of inputs/stimuli.   In particular, we can use  $f_\Phi(\Phi_x\setminus \Phi_y)$  to discriminate $x$ from $y$. 
Thus if $f_\Phi$ is fully deployable, then $f_{\Phi}(\Phi_x\setminus \Phi_y) = 0$ and   $f_{\Phi}(\Phi_y\setminus \Phi_x) = 0$ whenever $x\pind y$ or equivalently $
f_{\Phi}(\Phi_x\cap \Phi_y) = f(x) \mbox{ if $x\pind y$}$. 

We will call a set $D\subset \X$  \newterm{perceptually regular} if 
$[x]_{\sorites}\subset D$ for every $x\in D$. The classes in $R_c = \{x : \labelR{x}=c\}\subset \X$, where $R$ is a perceptually regular classifier are perceptually regular subsets.  Thus from now on we will use we will use the shorter name \newterm{regular class} instead of  a  perceptually regular subset. 

The \newterm{structural entropy} of the perceptually regular set $D$ is defined as: 
\begin{equation}\label{eq:general structural entropy}
H_{\sorites} (D) = -\frac{1}{\mu(D)}\mathop{\int}_{y\in D}\log\left(\frac{\mu([y]_{\sorites})}{\mu(D)}\right) \mbox{ and} 
\end{equation}
the\newterm{ index of coincidence} is
\begin{equation}\label{eq:IC}
\mbox{IC}_{\sorites} (D) =\frac{1}{\mu(D)^2}\mathop{\int}_{y\in D} \mu([y]_{\sorites}).
\end{equation}
The total/cumulative importance (salience) of the inputs in a regular class $D\subset \X$ and the expected affinity/similarity between pairs of inputs in $D$ can be defined and exploited if we have access to a well behaved (integrable)  importance scale $f_\Phi : \Ups \rightarrow [0,+\infty)$ and similarity scale $s : \X\times\X \rightarrow \R$:
\[
I_{\Phi}(D) = \int_{D} f_{\Phi}(\Phi_y) \mbox{ \newterm{the importance of} $D$}, 
\]
\[
{\cal R}_{\Phi}(D) = \frac{1}{\mu(D)}\int_{D}\int_{D} s(x,y) = \frac{1}{\mu(D)}\int_{D} P(x,D) \mbox{ \newterm{the expected affinity}.}
\]
The expected affinity defined above involves an iterated integral and is essentially the resemblance attribute defined in \cite{tversky1977features}. The following special case provides a particularly useful insight into the nature of prototypes and the possible internal  structure of categories as discussed in \cite{rosch1973internal}. \\

\myeg{proto} If $s$ is a contrast similarity \cite{tversky1977features} such that  $f_\Phi(\Phi_x) = s(x, x)/\theta$ is fully deployable,  $\theta\in (0,+\infty)$, and furthermore,  $f_\Phi(\Phi_x \cap \Phi_y) = f_\Phi(\Phi_x)$ for every pair $x\sorites y$ and $\Phi_x \cap  \Phi_y = \emptyset$ if $x\not\sorites y$, then for every regular class $D\subset \X$, and every $x\in D$ we get  
\begin{equation}\label{eq:proto_affinity}
P(x, D) = (\alpha + \beta + \theta)\mathop{\mu}(D)\left(\frac{\mu([x]_{\sorites})}{\mu(D)} - \frac{\alpha}{\alpha + \beta + \theta}\right)f_{\Phi}(\Phi_x) - \beta I_{\Phi}(D), 
\end{equation}
where $\alpha$, $\beta$, and $\theta$ are non-negative constants. 
In particular, if $D$ is a finite union of equivalence classes $\zeta_j\in \xmod$, then both prototypes and fringe elements exist. Furthermore, if $x$ is prototype/fringe then so are all stimuli in its component $\zeta = [x]_{\sorites}$.  Similar statements hold for M-core and $\tau$-fringe elements. 

 If the class $D$ is fixed, then $\displaystyle{\frac{\mu\left([x]_{\sorites}\right)}{\mu(D)}}$ is just the probability to encounter (and learn) a set of stimuli, and $f_{\Phi}(\Phi_x)$ can be interpreted as the level of prominence. So the optimization process involves learning prominent examples that can be encountered reasonably often. Thus in this case the prototype does not fall in either of the two main branches of prototypes, that is inputs that represent the central tendency in the regular class vs. prototypes as highly "representative exemplar(s)  of a category", see \cite{estes1994classification}, page 52. 

More generally, when the similarity scale is bounded, for example, this is true for real similarity measures deployable by humans, then as predicted by modern prototype theory perceptually regular subsets $D\subset \X$ corresponding to real (natural) categories created and analyzed by real intelligent agents consists of core elements (possibly M-core), a layer of fringe (possibly $\tau$-fringe) elements, and layers of elements of various levels of intermediate affinity with $D$. 
\section{Quantifying \dopp\ Vulnerability, \dopps\ attacks, fooling rate bound}\label{sec:aa}
By definition if a classifier $\classif{R}{m}$ is not regular, then it is vulnerable to adversarial \dopps\ attacks, that is, for some input $x$ there exists $a(x)\pind x$ and such that $\labelR{x}\neq \labelR{a(x)}$. In particular, we say that $R$ is \newterm{conceptually ambiguous}  at $x$ and we call the set 
\begin{equation}\label{eq:aiaia}
{\cal \textgreek{A}}(R)=\{x\in \X : \exists y\pind x\mbox{ and }\labelR{x}\neq\labelR{y}\}
\end{equation}
 the \newterm{region of conceptual ambiguity}. When $(X,\mu)$ is a probability measure space and $\mu(\twins{x}) > 0$, we use  the probability distribution  of labels at $x$:   
\begin{equation}\label{eq:pd_labels}
 p_j(x) =\frac{\mu\left(R_j\bigcap \twins{x}\right)}{\mu(\twins{x})},\hspace{1em} j=1,\ldots, m
\end{equation}
and the \newterm{conceptual entropy} of $R$  at $x$ defined as 
 \begin{equation}\label{eq:conceptual entropy at x}
 H_{R}(x) = -\sum_{j=1}^{m} p_j(x)\log(p_j(x))
 \end{equation}
 to detect whether $R$ is \CA\  at $x$ (i.e.,  $H_{R}(x) > 0 $), and to quantify the  likelihoods of various adversarial \dopp\ attacks. 
 
\begin{definition}\label{def:ADA}
Let  $R$ be a classifier and  $\hat{a}:\X\rightarrow \X$, we will call the inner measure of the set $\{x: \labelR{\hat{a}(x)}\neq \labelR{x}\}$ the $R$-\newterm{fooling rate} of the mapping $\hat{a}$, and we will denote it by $F_{R}(\hat{a}) $. Therefore 
\[
F_{R}(\hat{a}) = \mu_{*}\left(\{x: \labelR{\hat{a}(x)}\neq \labelR{x}\}\right).
\]

A mapping  $\hat{a}:\X\rightarrow \X$ is called an \newterm{adversarial \dopp\ attack} to a classifier $R$ if and only if $\hat{a}(x) \pind x$, $\forall x\in \X$,  and the $R$-fooling rate $F_{R}(\hat{a})$ is positive.  
\end{definition}

The set  $\{x: \labelR{\hat{a}(x)}\neq \labelR{x}\}$ is a subset of the region  of conceptual ambiguity of $R$, which yields an upper bound on the $R$-fooling rate
\begin{equation}\label{eq:FR_bound}
F_{R}(\hat{a}) \leq \mu_{*}(A(R)).
\end{equation}
In specific scenarios it is possible to get an upper bound on the  size, possibly the  outer measure, of $A(R)$ which in turn shows that the $R$-fooling rates are bounded away from one. See Example \ref{eg:Weber_R_plus_fool} below. 

\refstepcounter{e_counter}  \label{eg:Weber_R_plus_fool} \noindent {\bf Example  \ref{eg:Weber_R_plus_fool}:} Let $\mu(A) = \frac{2\sqrt{\pi}}{\pi}\int\limits_{A} e^{-t^2}\,d\!\,t$ be the probability measure on  $\X=(0, +\infty)$  and let the indiscriminability relation on $\X$ be defined by the covering   $\mathfrak{D}_{\textgreek{ad}} = \left\{\twins{x} = 
   \left(x/w, xw\right)\right\}_{x\in \X}$, where $w>1$ is a fixed constant. Let $\epsilon >0$ and let $R(\epsilon)$ be the linear classifier defined by 
   \begin{equation} \label{eq:binary}
\mbox{label}_{R(\varepsilon)}(x)= \begin{cases}
 1, &  0< x < \epsilon  \\
2, & x\geq\epsilon.
   \end{cases}
\end{equation}
 The conceptual entropy of $H_{R(\epsilon)}(x)$ is positive if and only if $x\in (\epsilon/w, w\epsilon)$. In particular, the points  outside the region of conceptual ambiguity  $\X\setminus  [\epsilon/w, w\epsilon]$ are not vulnerable to adversarial \dopps\ attacks. The conceptual entropy  achieves its global maximum $H_{R(\epsilon)}(x_{*}) = 1/2$ at a $x_* = x_*(\epsilon, w)$. It is equal to zero on $(0,\epsilon/w]\cup [\epsilon w, +\infty)$, and  increases monotonically  on $[\epsilon/w, x_*]$,  and then   decreases monotonically  on $[x_*, \epsilon w]$. The vulnerability to an adversarial \dopp\  attack is maximized at the point $x_*$. The  measure of the region of ambiguity $\mathop{\textgreek{A}}(R(\epsilon))$ is $\mathop{\mu}(\mathop{\textgreek{A}}(R(\epsilon)) = \erf{w\epsilon} - \erf{\epsilon/w} < 1$. The $R(\epsilon)$-fooling rate $F_{R(\epsilon)}(\hat{a}) \leq \erf{w\epsilon} - \erf{\epsilon/w} < 1$ of  an adversarial \dopp\ attack $\hat{a}$ is safely bounded away from $1$. 

Adversarial \dopp\  attacks are distinct from the adversarial attacks studied to date. The  universal adversarial attacks, \cite{moosavi2017universal},  can achieve fooling rates as close to one as one desires. As illustrated above, adversarial \dopp\  attacks may not be able to reach fooling rates that are too high. On the other hand in some cases, the optimal fooling rate of one can be achieved. 
\begin{obs}\label{obs:SADA}
If $R$ is conceptually ambiguous at every $x\in \X$, e.g., when  $H_R(x) > 0$ for every $x\in \X$, then there exists an adversarial \dopp\ attack with $R$-fooling rate equal to one. 
\end{obs}
Indeed, if $R$ is conceptually ambiguous at every $x\in \X$, then $\{y \in \twins{x}: \labelR{y}\neq \labelR{x}\} \neq \emptyset$, for every $x$, and therefore the axiom of choice implies that there exists a map $\hat{a}:\X\rightarrow \X$ such that $\hat{a}(x)\in \{y \in \twins{x}: \labelR{y}\neq \labelR{x}\}$, for every $x\in X$. It turns out that in practice there may be many classifiers that are conceptually ambiguous at every input. 

\refstepcounter{e_counter}  \label{eg:FR1} \noindent {\bf Example  \ref{eg:FR1}:} Consider the case when $\pind$ is transitive, i.e., $\pind$
equals its transitive closure  $\sorites$, e.g, when $\X$ is equipped with a Grassman structure as in \cite{krantz1975color}, and there exist at least two different equivalence classes. Thus binary classification is a well defined problem. There exist at least $\prod\limits_{\zeta \in \xmod}\left\{U\subset\zeta: 0 < \mu(U) < \mu(\zeta)\right\}$ worth of fully-populated binary classifiers which are conceptually ambiguous at every input $x\in \X$. 
 
The last example and Observation \ref{obs:SADA} show that even high accuracy classifiers can be vulnerable to adversarial \dopp\ attacks with fooling rate equal to one.  

We conclude this section with a warning that popular methods to deal with unseen data, including marking missing data and imputation, may introduce conceptual ambiguity.   For example, if a model is trained on a data set $T\subset \X$ that includes only parts of some elementary sets, then adding a  class label NA to label unseen data can compromise  adversarial \dopp\ robustness.  Indeed, let $S$ be a  nonempty set of  training data such that $S\subsetneq \zeta\in\xmod$. There exists, $x\in S$ such that $\twins{x}\setminus S\neq \emptyset$.  Every $z\in \twins{x}\setminus S$, labeled as NA, is an   adversarial \dopp\ of $x$.

\section{Discussion and Conclusions.}\label{sec:discussion}
A central focus of this paper is the adversarial  \dopps\ phenomenon, where  classifiers assign different labels to inputs that  humans cannot discriminate.  Until now, this phenomenon has  not been well understood, possibly due to the limitations of the distance-based analysis that has dominated the field. In the "absence of a distance measure that accurately captures the  perceptual  differences  between  a  source  and  adversarial  example  many researchers have decided to use the   $\ell_p$ distance",  \cite{hayes2018learning}. The available empirical observations and models - both perceptual and cognitive, including those based on just noticeable differences - provide no evidence that biologically plausible perceptual topologies are metric. This paper advances the understanding of context-related perceptual topologies  in input spaces, which are rarely metric. Our investigation shows that adversarial \dopps\ are very close to each other with respect to the context-relevant perceptual metric $d_w$, this metric is not a manifold metric and does not generate the perceptual topology.\footnote{The open metric ball $\mathring{B}^{w}_{1/2}(x)$ equals  $\{x\}$, 
for all inputs $x\in \X$. On the other hand, the finiteness of human observations and the hypothesis of bounded rationality suggest that biologically plausible perceptual topologies are not discrete.}  
This distinction highlights the shortcomings of traditional, purely manifold metric-based  representations and analysis of perceptual spaces.

The machine learning community has expended  significant efforts aimed to build adversarially robust classifiers. This may be  a march towards a bridge too far. Philosophers, experimental psychologists, and linguists, are well aware that many classification  problems are not well defined due to perceptual ambiguities. Any fully populated classifier for a classification problem that is not well defined is doomed to be a victim of the adversarial \dopps\ phenomenon. Our results reveal the structure of adversarial \dopp-robust classifiers,  regular classifiers,  and criteria and methods to establish whether a classification problem is well defined or not. The new understanding of the structure of regular classifiers, the analysis of zones of ambiguity, and the methods to measure and bound the fooling rates of adversarial \dopp\ attacks provide guidance  on how to design adversarially robust training to improve classifiers that are not regular.  In addition to revealing the impossibility to use accuracy-robustness trade-offs in many scenarios, including robustifying hyper-sensitive classifiers, our analysis indicates that marking unseen data can jeopardize robustness if the training data contains only a proper subset of an elementary set. 

We explore feature representations, the related concept of indiscernibility introduced by Leibniz,  and their connection to indiscriminability. This investigation reveals the nature of class prototypes and fringe inputs,  and how the size of a discriminative feature representation can be used to determine whether a classification problem is not well defined.  Indiscernibility and indiscriminability,  are often conflated in the machine learning literature. Elucidating the distinction between them is vital for understanding the limitations of current classifiers and addressing the shortcomings in their design.

Our discussion of the \dopps\ phenomenon brings to light a significant divergence between  human perception and artificial neural network models, including free-forward models, RNN models and ResNet. The  indiscriminability relations of  these artificial neural network models, studied in \cite{feather2019metamers}, are  transitive\footnote{See Part \ref{sect:ANN} in the Appendix.}  while it  is well accepted that in many  contexts the human indiscriminability relation is not transitive.

The results and insights gained from this investigation point to concrete warnings and actionable steps for improving the training and testing of classifiers.

% 
%Indiscriminability is guide to identity \cite{williamson1990identity} As discussed in  Section \ref{subs:leibniz} indiscernability is a guide to indiscriminability.  
\newpage

\bibliographystyle{alpha}
\bibliography{ms}

\newpage
%\clearpage
\pagenumbering{arabic}
\thispagestyle{plain}
\renewcommand{\shorttitle}{\textit{Appendix}}
\renewcommand{\headeright}{Proofs and Examples}
\appendix{\begin{center} \bf \large Golyadkin's Torment: Appendix \\ Odds, Ends, Proofs, and Examples\end{center}}
\section{Relations, Contrast, Coverings, Topology}\label{sect:tolerance}
Every reflexive binary relation (RBR) $\approx$ on $\X$ defines a  covering ${\displaystyle \left\{\mathfrak{g}(x) = \{y: x\approx y\}\right\}_{x\in \X}}$, of $\X$ and vice versa one can define reflexive binary relationships through coverings of $\X$.

\begin{lem}\label{lem:coverings} {\bf (a)\ } 
A covering $\mathfrak{G}=\{\mathfrak{g}(x)\}_{x\in \X}$, s.t., $x\in \mathfrak{g}(x), \forall x\in \X$ defines a unique RBR, 
\[
\approx_{\mathfrak{G}} = \left\{
(x, y): y\in \mathfrak{g}(x)
\right\}_{x\in\X}\subset\X\times\X; \footnote{We will use the  notation $\displaystyle{x\approx_{\mathfrak{G}}y, \mbox{ whenever }(x,y)\in \approx_{\mathfrak{G}}}$.}
\]
{\bf (b)\ }The RBR $\approx_{\mathfrak{G}}$ is symmetric if and only if $y\in \mathfrak{g}(x)$ whenever $x\in \mathfrak{g}(y)$; {\bf (c)\ } The RBR $\approx_{\mathfrak{G}}$ is transitive if and only if $\mathfrak{g}(y) \subset \mathfrak{g}(x)$ for all  $\displaystyle{(x,y)\in \approx_{\mathfrak{G}}}$.
\end{lem}
{\bf Proof of Lemma}.  
\noindent Parts {\bf (a)}  and {\bf (b)}  are trivial. 
If $\approx_{\mathfrak{G}}$ is transitive, then let  $x\approx_{\mathfrak{G}} y$. Since $\approx_{\mathfrak{G}}$ is transitive, every $z\in \mathfrak{g}(y)$, i.e.,  $y \approx_{\mathfrak{G}} z$, satisfies $x\approx_{\mathfrak{G}} z$, i.e., $z\in \mathfrak{g}(x)$.This proves the necessity in part {\bf (c)}. 
On the other hand, suppose that $\mathfrak{g}(y) \subset \mathfrak{g}(x)$ for all  $\displaystyle{(x,y)\in \approx_{\mathfrak{G}}}$, and so $\displaystyle{(x,y)\in \approx_{\mathfrak{G}}}$ and $\displaystyle{(y,z)\in \approx_{\mathfrak{G}}}$ imply $z\in \mathfrak{g}(z) \subset\mathfrak{g}(y) \subset \mathfrak{g}(x)$, which proves the sufficiency in part  {\bf (c)}.\\
\qed \\
Arguing by symmetry proves  the following 
\begin{obs}\label{obs:trans}
A symmetric RBR is transitive if and only if the canonical covering ${\displaystyle \left\{\mathfrak{g}(x) = \{y: y\approx x\}\right\}_{x\in \X}}$ satisfies  $\mathfrak{g}(x) = \mathfrak{g}(y)$ whenever $y\in \mathfrak{g}(x)$.
\end{obs}

Every RBR on a space $\X$ defines a canonical topology $\tau_\approx$  on $\X$: 
\begin{definition}\label{def:R_topo}
Let $\approx$ be a  RBR on a space $\X$, and let $\mathfrak{G}=\{\mathfrak{g}(x)\}_{x\in \X}$, s.t., $x\in \mathfrak{g}(x), \forall x\in \X$ be the  covering generated by $\approx$, the topology generated by the sub-basis $\mathfrak{G}$ is called the \newterm{canonical topology} generated by the RBR $\approx$ and is denoted by $\tau_\approx$. If the RBR is symmetric, the canonical topology is called \newterm{tolerable topology}.
\end{definition}
Tolerable/Tolerance topologies and some applications have been studied in \cite{poston1971fuzzy,hovsepian1992metalogical}, and \cite{peters2012tolerance}.

\subsection{Feature Representations and Indiscriminability}\label{subs:schreider}

{\bf Proof of Observation \ref{obs:law of indiscernibles}, Section \ref{subs:leibniz}.} \\
Let $x\pind y \pind z$, then under the assumption that $u\pind v$ implies  $\Phi_u = \Phi_v$, we get $\Phi_x = \Phi_y = \Phi_z$. The feature representation \LLP\ and so from $\Phi_x =  \Phi_z$ we get $x\pind z$. \qed

While indiscernibility has been discussed extensively, we believe that   active discrimination  of feature representations  is equally important biologically and epistemologically.  We will say that $x$ and $y$ are \newterm{actively indiscernible}, in a given context, if the subject is not able to activate the relevant knowledge that the features attributed to $x$ and the features attributed to $y$  are distinct. We will use the notation \newterm{$\Phi_x \pind \Phi_y$} to denote that $x$ and $y$ are actively indiscernible. 

Indiscernibility does imply active indiscriminibality, i.e, $(\Phi_x = \Phi_y)\implies (\Phi_x \pind \Phi_y)$. The former is a transitive relation on $\X$ but the later is a reflexive and symmetric relationship that may or may not be transitive. Many researchers have studied cases in which $(\Phi_x \pind \Phi_y)\implies (x \pind y)$, see for example  \cite{peters2007near,henry2011near,peters2012tolerance}. Alternatively, it is possible that there exist biologically plausible contexts where $(x \pind y) \implies (\Phi_x \pind \Phi_y)$. 

{\bf Proof of Theorem \ref{th:dfr exits}, Section \ref{subs:leibniz}.} \\
Let $\rgraph$ be the simple graph whose vertexes are  the points in $\X$ and the set of edges is $E_{\approx} = \left\{\{x,y\}, x\approx y\right\}$. Denote by $\rcliques$ the nonempty cliques in the graph $\rgraph$, and for every $x\in \X$ let $\rcliques_x = \{\varkappa\in \rcliques: x\in \varkappa\}$.  We define the feature space $\Phi$ to be the nonempty subsets of $\rcliques$ and the feature representation to be 
\[
\Phi_x = \rcliques_x, \forall x\in \X.
\] 
For every pair $x\approx y$,  $\Phi_x\cap \Phi_y\neq \emptyset$ since the clique $\{x, y\}\in \rcliques_x\cap \rcliques_y$. Vice versa, if  $\Phi_x\cap \Phi_y\neq \emptyset$, then $x, y$ belong to some clique $\kappa\in \rcliques$ and so $x\pind y$. \qed \\

\subsection{Removing Hypothetical and Semantically Synonymous Features.}\label{ref} 
 \Dfrs\ are very useful to model and study indiscrimination and categorization. However, the few explicit examples, including the maximal cliques proposed in \cite{schreider1975equality}, may be too  large to be biologically plausible. It is possible that some of the apparent bloat is due to the inclusion of hypothetical (non-attributable) and redundant features. 
 
Let $\Phi$ be a space of features,  and let $\{\Phi_x \subset \Phi\}_{x\in \X}$ be a context dependent feature representation of the inputs $x\in \X$. The \newterm{semantic cluster} sharing a feature $\xi\in \Phi$ is the context-dependent cluster of inputs: 
\begin{equation}\label{eq:s_cluster}
\clu{\xi} = \{x : \xi\in \Phi_{x}.\}
\end{equation}
The feature $\xi$ is hypothetical  in a given context if $\clu{\xi} = \emptyset$. 
The semantic clusters shared by the attributable features  define a new feature space $\clu{\Phi}$ and a new feature representation. Specifically, for every $x\in \X$ define $\clu{\Phi}_{x} = \{\clu{\xi}: \xi\in \Phi_{x}\}$ and the feature representation  $\{\clu{\Phi}_x \subset\clu{\Phi} = \mathop{\cup}\limits_{x\in\X} \clu{\Phi}_{x}\}_{x\in \X}$. There is a bijective mapping between the new feature space $\clu{\Phi}$ and the quotient space $\Phi/\!\!\equiv$ where the equivalence relation $\equiv$ is defined by $\xi\equiv \eta \Longleftrightarrow \clu{\xi}=\clu{\eta}$. 
In a sense the new feature space and representation are smaller (there is an on-to mapping $\Phi\rightarrow \clu{\Phi}$). All features in $\clu{\Phi}$ are attributed and there are no  redundant,  semantically  synonymous features.  Furthermore, 
\begin{equation}\label{eq:incl}
\clu{\xi}=  \clu{\clu{\xi}}, \mbox{ for every attributed feature }\xi\in \Phi.
\end{equation}
Indeed, if $\xi$ is attributed feature, then $y\in \clu{\xi}\Longleftrightarrow \xi\in \Phi_y \Longleftrightarrow \clu{\xi}\in \clu{\Phi}_y \Longleftrightarrow y\in \clu{\clu{\xi}}$. 

Let  $\approx$ be a reflexive and symmetric binary relation on a set $\X$, and let  $\{\Phi_x \subset \Phi\}_{x\in \X}$ be a $\approx$- \newterm{\dfr}, i.e., the representation satisfies Condition \ref{eq:gen_dfr}. The  clusters of inputs sharing attributed features  have additional structure and define a new \dfr. 
\begin{obs} \label{th:properties of clx} Let  $\approx$ be a reflexive and symmetric binary relation on a set $\X$, and let  $\{\Phi_x \subset \Phi\}_{x\in \X}$ be a $\approx$-\dfr, i.e., the representation satisfies Condition \ref{eq:gen_dfr}.  
\begin{description}
\item{\bf (i.)\ \, }For every attributed  feature $\xi\in \Phi$, the cluster $\clu{\xi}$ is a clique in $\rgraph$ and $\clu{\xi}\in \rcliques_{x}$, for every $x\in \clu{\xi}$.
\item{\bf (ii.)\ \, } $\left\{\clu{\Phi}_x\right\}_{x\in \X}$ is a $\approx$-\dfr.
\end{description} 
\end{obs}
{\bf Proof of Observation \ref{th:properties of clx}} 
\begin{description}
\item{\bf (i.)\ \, }Let $y, z\in \clu{\xi}$, and so $\xi\in \Phi_{y}\cap \Phi_{z}$, but $\{\Phi_x \subset \Phi\}_{x\in \X}$ is a $\approx$-\dfr\ and so $y\approx z$. Thus $\clu{\xi}\in \rcliques_x\subset \rcliques$, for all $x\in \clu{\xi}$.
\item{\bf (ii.)\ \, }  If $x\approx y$, then there exists $\xi\in \Phi_{x}\cap \Phi_y$ and so $\clu{\xi}\in  \clu{\Phi}_{x}\cap \clu{\Phi}_y$, and clearly $\clu{\Phi}_{x}\cap \clu{\Phi}_y \neq \emptyset$. 

On the other hand,  every feature $\xi\in \Phi$, such that $\clu{\xi}\in \clu{\Phi}_{x}\cap \clu{\Phi}_y \neq \emptyset$, belongs to $\Phi_{x}\cap \Phi_y$, and so $ \clu{\Phi}_{x}\cap \clu{\Phi}_y \neq \emptyset \implies \Phi_{x}\cap \Phi_y \neq \emptyset \implies x\approx y$.
\end{description} 

\qed

\subsection{Finite Discriminative Feature Representations}\label{subs:app:finite}
\begin{obs}\label{ob: finite dfr}
Let $\approx$ be symmetric and reflexive binary relation on $\X$, $\Phi$ a finite set of attributed features. If there exists an $\approx$-\dfr\  $\{\Phi_x\subset \Phi\}_{x\in \X}$, then  for every fully populated classifier $\classif{R}{m}$ with more labels than the total number of attributed features $\#\Phi$, $m>\#\Phi$, there exist $x\approx y$ such that $\labelR{x}\neq \labelR{y}$. 
\end{obs}
{\bf Proof of Observation \ref{ob: finite dfr}} \\
Let $\Phi_{(j)}$ be the collection of features attributed to inputs whose labels equal $j = 1,\ldots, m$. Specifically,
\[
\Phi_{(j)} = \{\xi\in \Phi : \xi\in \Phi_x \mbox{ for some $x\in R_j\subset \X$}\}.
\]
Denote by $\#_j$ the number of elements  in $\Phi_{(j)}$, i.e., $\#_j = \#\Phi_{(j)}$.  The classifier is fully populated and the feature representation is $\approx$-\dfr\ and so 
\begin{equation}\label{obs: number bound}
\#_j \geq 1,\hspace{1em}  j=1,\ldots, m.
\end{equation}
The sets of attributed features $\Phi_{(j)}$ cannot be disjoint. Indeed, $\Phi_{(i)}\cap\Phi_{(j)} = \emptyset$, if $i\neq j$ together with Inequality \ref{obs: number bound}would lead to the contradiction 
\[
m>\#\Phi = \sum_{j=1}^{m} \#_j \geq m. 
\]
Thus there exits a feature $\xi \in \Phi_{(i)}\cap \Phi_{(j)}$ for some labels $i\neq j$. And hence there exist two inputs $x\in R_i$ and $y\in R_j$, i.e., $\labelR{x} = i\neq j = \labelR{y}$, such that $\xi\in \Phi_x \cap \Phi_y$. The feature representation is $\approx$-discriminative and so $x\approx y$. \\
\qed
\subsection{Leibniz  and {\it what is it to be an attribute\ldots}\textnormal{\cite{von1902leibniz}, Section VIII.}} In his 1686 {\em Discourse on Metaphysics}, \cite{von1902leibniz},  Leibniz discusses the understanding of substances by a perfect being.   The following example  summarizes and builds on the ideas expounded by Leibniz.   

\refstepcounter{e_counter}   \label{eg:metaVIII} \noindent {\bf Example  \ref{eg:metaVIII}:} Let $\Phi$ be the collection of all predicates and assume that there exists a sentient agent that has the capacity to build a feature representation   where $\Phi_x$ consists of all predicates $P$ such that $P(x) = \mbox{T}$ in the given context. We will call this the \newterm{Metaphysical Representation}. This feature representation \LLP.

Indeed, assuming  the existence of a pair of inputs $x$ and $y$ for which   $\Phi_x = \Phi_y$ but such that the sentient agents do discriminate $x$ from $y$ leads to a contradiction.  To arrive to the contradiction let $P$ be the predicate ``Does not belong to $\twins{y}$". $P\in \Phi_{x}$ since, in the fixed context, the agents do discriminate $x$ from $y$. However,  $P\not\in \Phi_y$ since $y\in\twins{y}, \forall y\in\X$. Thus $\Phi_x\neq \Phi_y$ which contradicts the assumption that $\Phi_x = \Phi_y$. 

\section{Sorites, Ill-posed Classification}\label{sect:sorites}
Sorites chains and the related paradoxes are deeply related to indiscriminability. They have been studied and argued since at least the 4th century BCE.\footnote{At least since Eubulides of Miletus formulated the Heap Paradox.} Every pair of adversarial \dopps\ $x\pind y$ such that $\labelR{x}\neq \labelR{y}$ is a sorites chain. On the other hand, every sorites chain  $ x_1\pind x_2\pind\cdots \pind x_n$ such that $\labelR{x_1}\neq \labelR{x_n}$ for some classifier $R$ must contain a pair of adversarial \dopps. Indeed, 
\begin{lem} \label{lemma:chain} Let  $R$ be classifier  with  labeling function $\mbox{label}_R : X\rightarrow \{1, 2, \ldots,m\}$. If there exists a chain of \dopps\ $ x_1\pind x_2\pind\cdots \pind x_n$ whose initial and final samples are assigned different labels  $ \labelR{x_1}\neq\labelR{x_n}$ by $R$  then there exists a pair of \adv\ \dopps\  
$ x_{i}\pind x_{i+1}$, $ \labelR{x_i}\neq\labelR{x_{i+1}}$, where $i\in \{1, \ldots,n-1\}$. In fact  $i$ can be chosen so that 
$ \labelR{x_1} = \labelR{x_{i}}$   and/or  $ \labelR{x_{i+1}}=\labelR{x_n}$.
\end{lem}
{\bf Proof of Lemma}.  
The short proof of the lemma is constructive. There are two types of constructions that might be used to produce the \adv\ \dopps, \ the ``first encounter pair" and the ``last encounter".  The last encounter construction is based on identifying the last link in the chain that has the same label as $x_1$. Set
\[
i=\max\{j\in  \{1, 2, \ldots,m\} \, :\, \labelR{x_1} = \labelR{x_{j}} \}. 
\]
Then  $ \labelR{x_i} = \labelR{x_{1}}$ and  $ \labelR{x_i} = \labelR{x_{1}}\neq \labelR{x_{i+1}}$. Thus   $ x_{i}\pind x_{i+1}$ is a  pair of \adv\ \dopps. The first encounter construction is based on identifying the first link in the chain whose label is different from $\labelR{x_1}$. Indeed, let  
\[
i=\min\{j\in  \{1, 2, \ldots,m\} \, :\, \labelR{x_1} \neq \labelR{x_{j}} \}.   
\]
Then $ \labelR{x_{i-1}} = \labelR{x_{1}}$ and $ \labelR{x_{i-1}}=  \labelR{x_{1}} \neq\labelR{x_{i}}$. Thus  $ x_{i-1}\pind x_{i}$, are a pair of  \adv\ \dopps. To complete the proof of the lemma reverse the order of elements in the \dopp\ chain.  
\\
\qed 

\begin{lem}\label{lemma:web}  Let $\X$ be the  closed bounded interval $[a, b]\subset (0, +\infty)$. Suppose that Weber's law holds  and let $k$ be the Weber constant. Let $w = 1+k$,  and  
\begin{equation*} 
\twins{x} = \begin{cases}
  \left[a, xw\right), & a\leq x <  aw \\
   \left(x/w, xw\right), & aw\leq x \leq  b/w \\
     \left(x/w, b\right],  & b/w < x \leq  b,
 \end{cases}
\end{equation*} 
then $\xmod$ is a singleton. 
\end{lem}
{\bf Proof of Lemma}.
Indeed, an argument by induction shows that every  two inputs $x\in \X$ and $y\in \X$ can be ``connected" by a finite chain of \dopps\  $x\pind x_1\pind x_2\pind\cdots \pind x_m\pind y$. The induction will be on  $\mbox{jump}(x;y)=\displaystyle{\min\{l\in \N:\,\, y<w^lx\}}$  \newterm{the number of (perceptual) jumps}  one needs to get from $x$ to $y$.  Note that by the definition of the sub-basis  $\mbox{jump}(x;y)=1$ implies $x\pind y$.  

Let us assume that $\mbox{jump}(x;z)\leq n$ implies that $x$ can be connected to $ z$ by a finite chain of \dopps, and suppose that $\mbox{jump}(x;y) = n+1$. 

\begin{figure}[h]
\begin{center}
%%\framebox[4.0in]{$\;$}
%\fbox{\rule[-.5cm]{0cm}{2cm} \rule[-.5cm]{4cm}{0cm}}
\includegraphics[width=12cm]{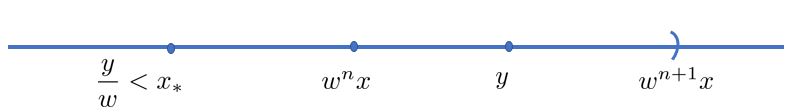}
\end{center}
\caption{ Linking  $x\pind x_1\pind x_2\pind\cdots \pind x_*$ with  $x_{*}\pind y$ to get the chain of \dopps\ $x\pind x_1\pind x_2\pind\cdots \pind x_*\pind y$  from $x$ to $y$. }  \label{fig:sorites_exist_proof}
\end{figure}

If  $w^{n}x < y < w^{n+1}x$, choose any $x_*$ such that  $y/w < x_* < w^nx$  (see  Figure \ref{fig:sorites_exist_proof}). 
 
Since $\mbox{jump}(x;x_*) =  n$, the  induction assumption implies that there exists a finite chain of \dopps\ $x\pind x_1\pind x_2\pind\cdots \pind x_*$. By construction $x_{*}\pind y$, and  so we can add this final link to obtain the finite chain $x\pind x_1\pind x_2\pind\cdots \pind x_*\pind y$ . 
\qed

Thus if $R$ is any classifier defined on $[a, b]$ such that $\labelR{x} \neq \labelR{y}$  for some $a\leq x, y \leq b$, then $x\sorites y$  and Lemma \ref{lemma:chain} implies that $R$ admits adversarial \dopps\ and hence cannot be regular. 

More generally, if  the transitive closure $\sorites$  of $\pind$  is trivial then every fully populated classifier with two or more classes admits \adv\ \dopps. In particular, let $R$ be a fully populated classifier  with  labeling function $\mbox{label}_R : X\twoheadrightarrow \{1, 2, \ldots,m\}$, then for every label $c$ there exist \adv\ \dopps\  $x(c)\pind x^{*}(c)\in \X$  such that $c=\labelR{x(c)}\neq\labelR{x^{*}(c)}$. 

\section{Adversarial Training May or May Not Work}\label{sect:adv}
It turns out that it is possible to use  appropriate ``adversarial training" to  improve accuracy as well as adversarial \dopps\ robustness. \\

{\bf Proof of Observation \ref{claim:miss_class_adv_dop}, Section \ref{subs:can}.} \\
If    $\displaystyle{\inf_{x\in\X} \mu(\twins{x}) > 0}$, then for every classifier 
$R = \{R_1,\ldots, R_m\}$ whose recall rates are sufficiently high so that  $\underline{\rho}>1-1/\bar{k}(\Omega)$. i.e.,
\begin{equation}\label{eq:high}
\left(1-\underline{\rho}\right)\bar{k}(\Omega)<1
\end{equation}
Then every   misclassified input $x$ is an adversarial \dopp. 

\noindent {\bf Proof}. 
Let $x$ be misclassified by $R$, that is $x\in R_j\cap\Omega_{i(x)}$, where $j\in \{1,\ldots,m\}$ and $j\neq i(x)$ and so 
\begin{equation}\label{eq:est miss class in omega}
\mu\left(\classof{\Omega}{x}\right)  -  \mu\left(\classof{\Omega}{x}\cap \classof{R}{x}\right) =  \sum_{s\neq i(x)} \mu\left(\classof{\Omega}{x}\cap R_s\right) \geq  \mu\left(\classof{\Omega}{x}\cap R_j\right)
\end{equation}
and so 
\[
1-\underline{\rho} \geq 1- \frac{ \mu\left(\classof{\Omega}{x}\cap \classof{R}{x}\right) }{ \mu\left(\classof{\Omega}{x}\right) } \geq \frac{\mu\left(\classof{\Omega}{x}\cap R_j\right) }{ \mu\left(\classof{\Omega}{x}\right) } 
\]

 The lower bound of the recall rates  $\left(1-\underline{\rho}\right)\bar{k}(\Omega)<1$ 
 and  $\twins{x}\subset \classof{\Omega}{x}$  yield the estimate
\begin{equation}\label{eq:positive bound}
1 > \left(1-\underline{\rho}\right)\bar{k}(\Omega) \geq  \frac{\mu\left(\classof{\Omega}{x}\right)}{\mu\left(\twins{x}\right)}\frac{\mu\left(\classof{\Omega}{x}\cap R_j\right) }{ \mu\left(\classof{\Omega}{x}\right) } \geq \frac{\mu\left(\twins{x}\cap R_j\right)}{\mu\left(\twins{x}\right) }.
\end{equation}
Hence the set of adversarial \dopps\  of $x$   has positive measure, i.e., $\mu\left(\twins{x}\setminus R_{j}\right) > 0$. \\
 \noindent \qed
 
However,  the following example shows this may be a behavior that is a side effect of the nature of robust training that targets data that is as close as possible to a decision boundary.  
% The later may be true if there are crisply defined threshold levels or measure zero category differentiation limens. 
\vskip 12pt

\myeg{sugit robust disciplina}  Let  $\X=\R$ be the real line and let the probability measure $\mu$ be the Gaussian with mean 0 and variance $\sigma^2 = 1/2$ . Suppose that Weber's law holds  and let $k>0$ be the Weber constant. Let $w = 1+k$,  and let 
\begin{equation} \label{eq:k line}
\twins{x} = \begin{cases}
   \left(wx, x/w\right), &  x <0  \\
   \{0\}, & x = 0 \\
   \left(x/w, xw\right), &  x >0 
   \end{cases}
\end{equation}
Then $\xmod = \left\{(-\infty, 0), \{0\}, (0, +\infty) \right\}$ and 
\begin{equation} \label{eq:mu dopp}
\mu(\twins{x}) = \begin{cases}
\displaystyle{\frac{ \left|  \erf{wx} -\erf{x/w}\right|}{2}}, &  x \neq 0  \\
   0, & x = 0 
   \end{cases}
\end{equation}
Consider the  regular model $\Omega$ two classes $\Omega_1 =  (-\infty, 0)$ and $\Omega_2 = [0, +\infty)$. Let $\epsilon > 0$ and let $R(\varepsilon)$ be a two label classifier such that 
\begin{equation} \label{eq:sugit binary}
\mbox{label}_{R(\varepsilon)}(x)= \begin{cases}
 1, &  x < \epsilon  \\
2, & x\geq\epsilon
   \end{cases}
\end{equation}
\begin{figure}[h]  
\begin{center}
%%\framebox[4.0in]{$\;$}
%\fbox{\rule[-.5cm]{0cm}{4cm} \rule[-.5cm]{4cm}{0cm}}
\includegraphics[width =10cm]{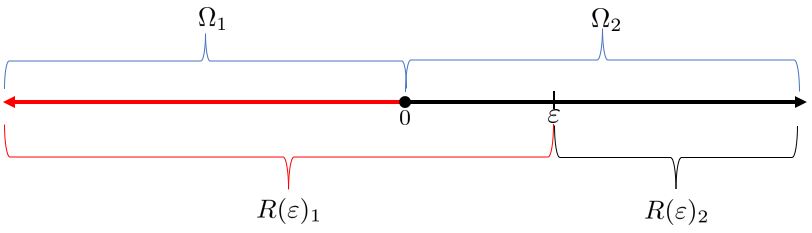}
\end{center}
\caption{$X=\R$, $\twins{x}$ defined in  (\ref{eq:k line}), $\Omega_1 =  (-\infty, 0)$ and $\Omega_2 = [0, +\infty)$, the classifier $R(\varepsilon)$ defined in (\ref{eq:sugit binary}).}
\label{fig:kline}
\end{figure}
Then the misclassified inputs are $[0,\epsilon)$, $\twins{\epsilon}$ is the collection of all adversarial \dopps\ and  the set of misclassified adversarial \dopps\ is $(\epsilon/w, \epsilon)$. 

\begin{figure}[h]
\begin{center}
%%\framebox[4.0in]{$\;$}
%\fbox{\rule[-.5cm]{0cm}{4cm} \rule[-.5cm]{4cm}{0cm}}
\includegraphics{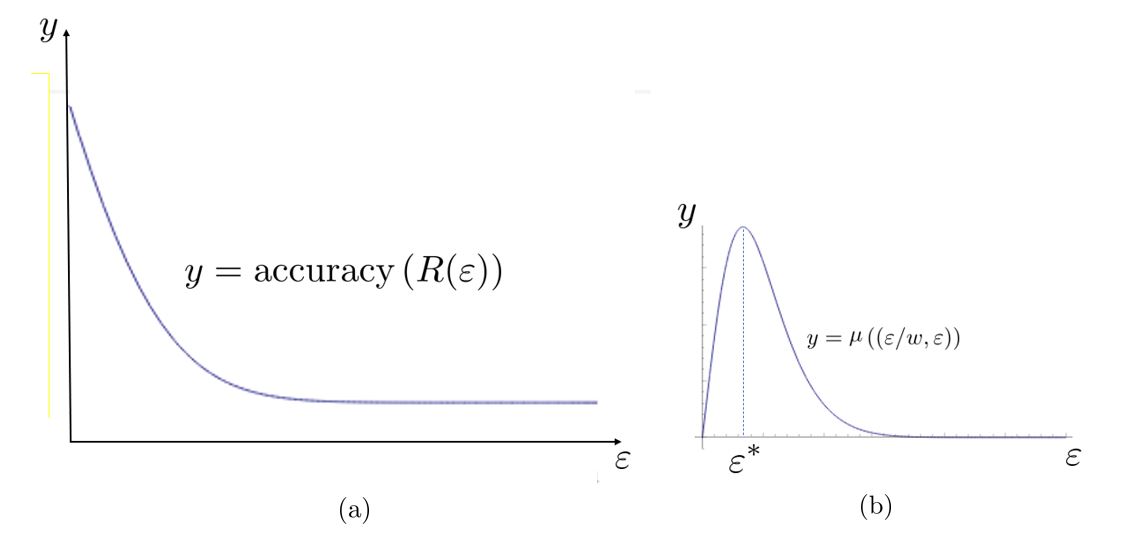}
\end{center}
\caption{(a) The accuracies of the linear classifiers  $R(\varepsilon)$ decrease as $\varepsilon$ increases. (b) The sizes of the sets of adversarial \dopps\ of the linear classifiers $R(\varepsilon)$ have a unique maximal value achieved at $\varepsilon^*$.  The graph of the sizes of the sets of adversarial \dopps\  is shown in this panel. To improve the readability the graph  is scaled up to match the scales of the accuracy graph.}  \label{fig:accuracy_vs_ robustness}
\end{figure}

Using the adversarial training \cite{papernot2016limitations}, will move the decision boundary towards zero thus improving the accuracy of the classifier (at best the new decision boundary  moves to $x=\epsilon/w$). The good news is that the "robustified" classifiers will converge to the perfect accuracy  classifier $\Omega$. See Figure \ref{fig:accuracy_vs_ robustness}(a)

%
%On the other hand clearly we can robustify against adversarial \dopps\ a classifier  with sufficiently low accuracy, lower than  the accuracy of the classifier $R(\varepsilon^*)$,  but we will have to give up accuracy.   
%\vskip 12pt

However, the function $\mu((\varepsilon/w,\varepsilon))$ has a unique global maximum on $[0,+\infty)$ achieved at $\varepsilon^*>0$, see Figure \ref{fig:accuracy_vs_ robustness}(b) . Thus if we apply robust training starting with a classifier s.t. $\varepsilon > \varepsilon^*$ then we will improve accuracy but gain adversarial \dopps\ (seemingly, we will be trading off  adversarial \dopp\ robustness to gain accuracy). However, if   $\varepsilon < \varepsilon^*$, robust training  will improve both accuracy and adversarial \dopp\ robustness (i.e.,  there will be no trade-off just gain across the front) as we move down towards the perfect accuracy classifier.

\section{The Laplace Operator $\plap$}\label{sect:laplace}
Assuming that the degree of a vertex $x$ in the graph $\pgraph$  defined by 
\begin{equation}\label{eq:pnode_degree}
\pdeg(x) = \mu(\twins{x})
\end{equation}
 is integrable and 
$\displaystyle{\inf_{x\in \X}}(\pdeg(x)) > 0$, then we define the \newterm{discrimination Laplace operator}  c.f. \cite{chung1996lectures}:
\begin{equation}\label{eq:plap}
\plap(f) (x)= f(x) - \frac{1}{\sqrt{\pdeg(x)}}\int_{\twins{x}}\frac{f}{\sqrt{\pdeg}}
\end{equation}
The kernel of $\plap$ is nontrivial since $\plap(\sqrt{\pdeg}) = 0$ and so $0$ is an eigenvalue of $\plap$. On the other hand, globally constant functions $f(x) \equiv c\in \R\setminus \{0\}$ are harmonic if and only if
\begin{equation}\label{eq:constant harmonic} 
\frac{1}{\sqrt{\pdeg(x)}}\int_{\twins{x}}\frac{1}{\sqrt{\pdeg}} = 1, \forall x\in\X.
\end{equation}
It is easy to show that  (\ref{eq:constant harmonic})  fails in  Example \ref{eg:Weber} and Example \ref{eg:Weber2}, where the probability measure $\mu$ is the uniform measure and so all nontrivial globally constant functions are not $\plap$ harmonic. 
On the other hand, if $\pind$ is transitive, then $\twins{x}=[x]_{\sim_{\sigma}}, \forall x\in \X$, and so $\plap = \slap$.

\section{Harmonic salience functions}\label{sect:features}
\noindent By definition, a salience scale$f_{\Phi}$ is perceptually regular iff $f_{\Phi}\left(\Phi_{x}\right)=f_{\Phi}\left(\Phi_{y}\right)$ whenever $x\sorites y$, i.e., whenever the corresponding salience function $f:\X\rightarrow \R$ is perceptually regular and hence harmonic with respect to $\slap$ ($\slap f = 0$). 

\section{ANN Discrimination}\label{sect:ANN}
The indiscriminability of inputs by  VGG-19, ResNet, and Inception-V3  has been studied by Feather et al., \cite{feather2019metamers}. Two inputs $x$ and $y$ are indiscriminable by these ANN models, $x\stackrel{\text{\tiny ANN}}{\approx} y$, iff they ``produce the same activations
in a model layer''. The relation $\stackrel{\text{\tiny ANN}}{\approx}$ is transitive. Indeed, let $x$ and $y$ produce the same activations at some level, then they produce the same activations in all subsequent levels. Thus if $x\stackrel{\text{\tiny ANN}}{\approx} y$ and $y\stackrel{\text{\tiny ANN}}{\approx} z$, then $x$ and $z$ produce the same activations at all sufficiently high levels, and therefore, $x\stackrel{\text{\tiny ANN}}{\approx} z$. The same argument does imply that $\stackrel{\text{\tiny ANN}}{\approx}$ is transitive for all free-forward models and for all recurrent neural network models.  
\end{document}